\documentclass[10pt,twocolumn,letterpaper]{article}

\usepackage{cvpr}
\usepackage{times}
\usepackage{epsfig}
\usepackage{graphicx}
\usepackage{amsmath}
\usepackage{amssymb}
\usepackage{xcolor}
\usepackage{authblk}

\cvprfinalcopy 
\begin{document}

\definecolor{olive}{RGB}{50,150,50}

\newif\ifdraft
\drafttrue

\ifdraft
\newcommand{\PF}[1]{{\color{red}{\bf pf: #1}}}
\newcommand{\pf}[1]{{\color{red} #1}}
\newcommand{\HR}[1]{{\color{blue}{\bf hr: #1}}}
\newcommand{\hr}[1]{{\color{blue} #1}}
\newcommand{\sg}[1]{{\color{orange}{#1}}}
\newcommand{\SG}[1]{{\color{orange}{\bf sg: #1}}}
\newcommand{\pr}[1]{{\color{green}{#1}}}
\newcommand{\PR}[1]{{\color{green}{\bf pr: #1}}}
\newcommand{\sL}[1]{{\color{green}{#1}}}
\newcommand{\SL}[1]{{\color{green}{\bf sl: #1}}}
\else
\newcommand{\PF}[1]{{\color{red}{}}}	
\newcommand{\pf}[1]{ #1 }
\newcommand{\HR}[1]{{\color{blue}{}}}
\newcommand{\hr}[1]{#1}%
\newcommand{\VC}[1]{{\color{green}{}}}
\newcommand{\ms}[1]{ #1 }
\newcommand{\MS}[1]{{\color{olive}{}}}
\newcommand{\NEW}[1]{#1}
\newcommand{\sg}[1]{{\color{orange}{#1}}}
\newcommand{\SG}[1]{{\color{orange}{\bf sg: #1}}}
\newcommand{\pr}[1]{{\color{green}{#1}}}
\newcommand{\PR}[1]{{\color{green}{\bf pr: #1}}}
\renewcommand{\dm}[1]{{\color{yellow}{#1}}}
\newcommand{\DM}[1]{{\color{yellow}{\bf pr: #1}}}
\fi

\newcommand{\comment}[1]{}

\newcommand{\datasetSize}{40,063}
\newcommand{\datasetSizeManual}{5,063}
\newcommand{\datasetSizeAutomatic}{29,000}
\newcommand{\datasetSizeActive}{6,000}
\newcommand{\datasetSizeTrain}{37,000}
\newcommand{\datasetSizeTest}{2,063}
\newcommand{\datasetSizeVal}{1,000}

\newcommand{\TODO}[1]{\textcolor{cyan}{#1}}

\newcommand{\ST}{\mathcal{T}}
\newcommand{\SST}{\mathcal{T}_S}

\newcommand{\R}{\mathbb{R}}
\newcommand{\Seg}{\mathbf{S}} 
\newcommand{\Latent}{\mathbf{L}}
\newcommand{\LatentG}{\Latent^{\text{3D}}} 
\newcommand{\LatentA}{\Latent^\text{app}} 
\newcommand{\LatentBG}{\mB} 

\newcommand{\dA}{{A}}
\newcommand{\dB}{{B}}

\newcommand{\parag}[1]{\vspace{-3mm}\paragraph{#1}}

\newcommand{\va}{\mathbf{a}}
\newcommand{\vb}{\mathbf{b}}
\newcommand{\vd}{\mathbf{d}}
\newcommand{\ve}{\mathbf{e}}
\newcommand{\vf}{\mathbf{f}}
\newcommand{\vg}{\mathbf{g}}
\newcommand{\vh}{\mathbf{h}}
\newcommand{\vi}{\mathbf{i}}
\newcommand{\vj}{\mathbf{j}}
\newcommand{\vk}{\mathbf{k}}
\newcommand{\vl}{\mathbf{l}}
\newcommand{\vm}{\mathbf{m}}
\newcommand{\vn}{\mathbf{n}}
\newcommand{\vo}{\mathbf{o}}
\newcommand{\vp}{\mathbf{p}}
\newcommand{\vq}{\mathbf{q}}
\newcommand{\vr}{\mathbf{r}}
\newcommand{\vt}{\mathbf{t}}
\newcommand{\vu}{\mathbf{u}}
\newcommand{\vv}{\mathbf{v}}
\newcommand{\vw}{\mathbf{w}}
\newcommand{\vx}{\mathbf{x}}
\newcommand{\vy}{\mathbf{y}}
\newcommand{\vz}{\mathbf{z}}

\newcommand{\mA}{\mathbf{A}}
\newcommand{\mB}{\mathbf{B}}
\newcommand{\mC}{\mathbf{C}}
\newcommand{\mD}{\mathbf{D}}
\newcommand{\mE}{\mathbf{E}}
\newcommand{\mF}{\mathbf{F}}
\newcommand{\mG}{\mathbf{G}}
\newcommand{\mH}{\mathbf{H}}
\newcommand{\mI}{\mathbf{I}}
\newcommand{\mJ}{\mathbf{J}}
\newcommand{\mK}{\mathbf{K}}
\newcommand{\mL}{\mathbf{L}}
\newcommand{\mM}{\mathbf{M}}
\newcommand{\mN}{\mathbf{N}}
\newcommand{\mO}{\mathbf{O}}
\newcommand{\mP}{\mathbf{P}}
\newcommand{\mQ}{\mathbf{Q}}
\newcommand{\mR}{\mathbf{R}}
\newcommand{\mS}{\mathbf{S}}
\newcommand{\mT}{\mathbf{T}}
\newcommand{\mU}{\mathbf{U}}
\newcommand{\mV}{\mathbf{V}}
\newcommand{\mW}{\mathbf{W}}
\newcommand{\mX}{\mathbf{X}}
\newcommand{\mY}{\mathbf{Y}}
\newcommand{\mZ}{\mathbf{Z}}

\newcommand{\cA}{\mathcal A}
\newcommand{\cB}{\mathcal B}
\newcommand{\cC}{\mathcal C}
\newcommand{\cD}{\mathcal D}
\newcommand{\cE}{\mathcal E}
\newcommand{\cF}{\mathcal F}
\newcommand{\cG}{\mathcal G}
\newcommand{\cH}{\mathcal H}
\newcommand{\cI}{\mathcal I}
\newcommand{\cJ}{\mathcal J}
\newcommand{\cK}{\mathcal K}
\newcommand{\cL}{\mathcal L}
\newcommand{\cM}{\mathcal M}
\newcommand{\cN}{\mathcal N}
\newcommand{\cO}{\mathcal O}
\newcommand{\cP}{\mathcal P}
\newcommand{\cQ}{\mathcal Q}
\newcommand{\cR}{\mathcal R}
\newcommand{\cS}{\mathcal S}
\newcommand{\cT}{\mathcal T}
\newcommand{\cU}{\mathcal U}
\newcommand{\cV}{\mathcal V}
\newcommand{\cW}{\mathcal W}
\newcommand{\cX}{\mathcal X}
\newcommand{\cY}{\mathcal Y}
\newcommand{\cZ}{\mathcal Z}

\title{Deformation-aware Unpaired Image Translation for Pose Estimation on Laboratory Animals} 

\author[1]{Siyuan Li}
\author[1,3]{Semih G\"{u}nel}
\author[1]{Mirela Ostrek}
\author[3]{Pavan Ramdya}
\author[1]{Pascal Fua}
\author[1,2]{Helge Rhodin}

\affil[1]{CVLAB, EPFL, Lausanne}
\affil[2]{Imager Lab, UBC, Vancouver}
\affil[3]{Neuroengineering Lab, EPFL, Lausanne}

\maketitle
\begin{abstract}
	
Our goal is to capture the pose of neuroscience model organisms, without using any manual supervision, to be able to study how neural circuits orchestrate behaviour.
Human pose estimation attains remarkable accuracy when trained on real or simulated datasets consisting of millions of frames.
However, for many applications simulated models are unrealistic and real training datasets with comprehensive annotations do not exist. 
We address this problem with a new sim2real domain transfer method.
Our key contribution is the explicit and independent modelling of appearance, shape and pose in an unpaired image translation framework.
Our model lets us train a pose estimator on the target domain by transferring readily available body keypoint locations from the source domain to generated target images.We compare our approach with existing domain transfer methods and demonstrate improved pose estimation accuracy on Drosophila melanogaster (fruit fly), Caenorhabditis elegans (worm) and Danio rerio (zebrafish), without requiring any manual annotation on the target domain and despite using simplistic off-the-shelf animal characters for simulation, or simple geometric shapes as models.  
Our new datasets, code and trained models will be published to support future neuroscientific studies.

\end{abstract}



\section{Introduction}

Deep learning-based pose estimation on images has evolved into a practical tool for a wide range of applications, as long as sufficiently large training databases are available.  However, in very specialized domains there are rarely large annotation databases. For example, neuroscientists need to accurately capture the poses of all the appendages of fruit flies, as pose dynamics are crucial for drawing inferences about how neural populations coordinate animal behavior. Publicly available databases for such studies are rare and current annotation techniques available to create such a database are tedious and time consuming, even when semi-automated. Given the existence of motion simulators, an apparently simple workaround would be to synthesize images of flies in various poses and use these synthetic images for training purposes. Although image generation algorithms can now generate very convincing {\it deepfakes}, existing image translation algorithms do not preserve pose geometrically when the gap between a synthetic source and a real target is large. This is critical to our application, as creating matching high-fidelity images would be time consuming. 

\begin{figure}[t]
	\centering
	\includegraphics[width=1.0\linewidth]{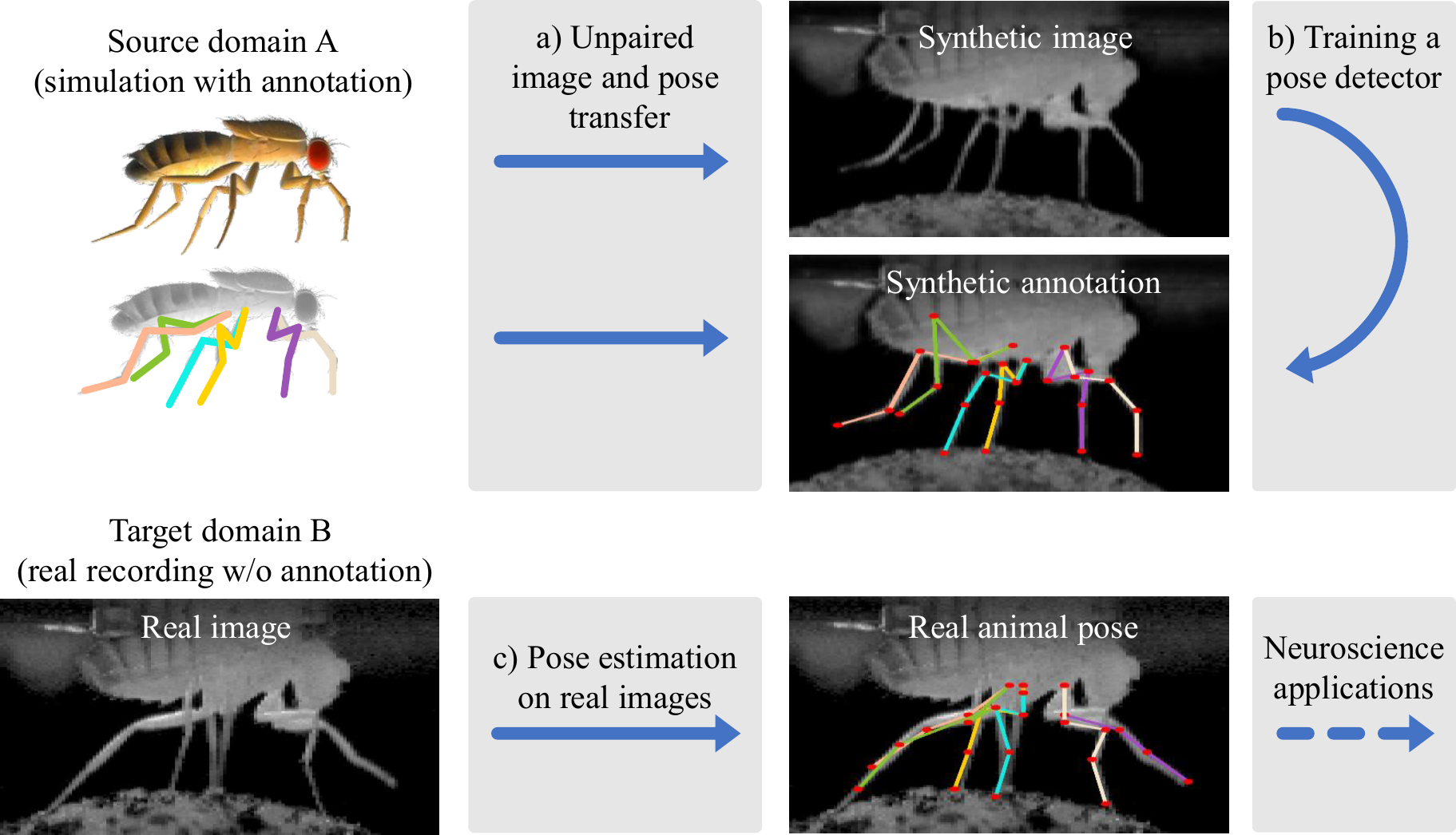}
	\caption{\textbf{Approach.}  Our most morphologically complex example is the six-legged \textit{Drosophila}: a) We transfer synthetic images and their keypoint annotations to realistically looking images using only unpaired examples of the two domains. b) Our method enables training of a pose detector that c) can be applied to real images for neuroscientific studies.}
	\label{fig:teaser}
\end{figure}

\begin{figure}[t]
\centering
		\includegraphics[width=0.8\linewidth]{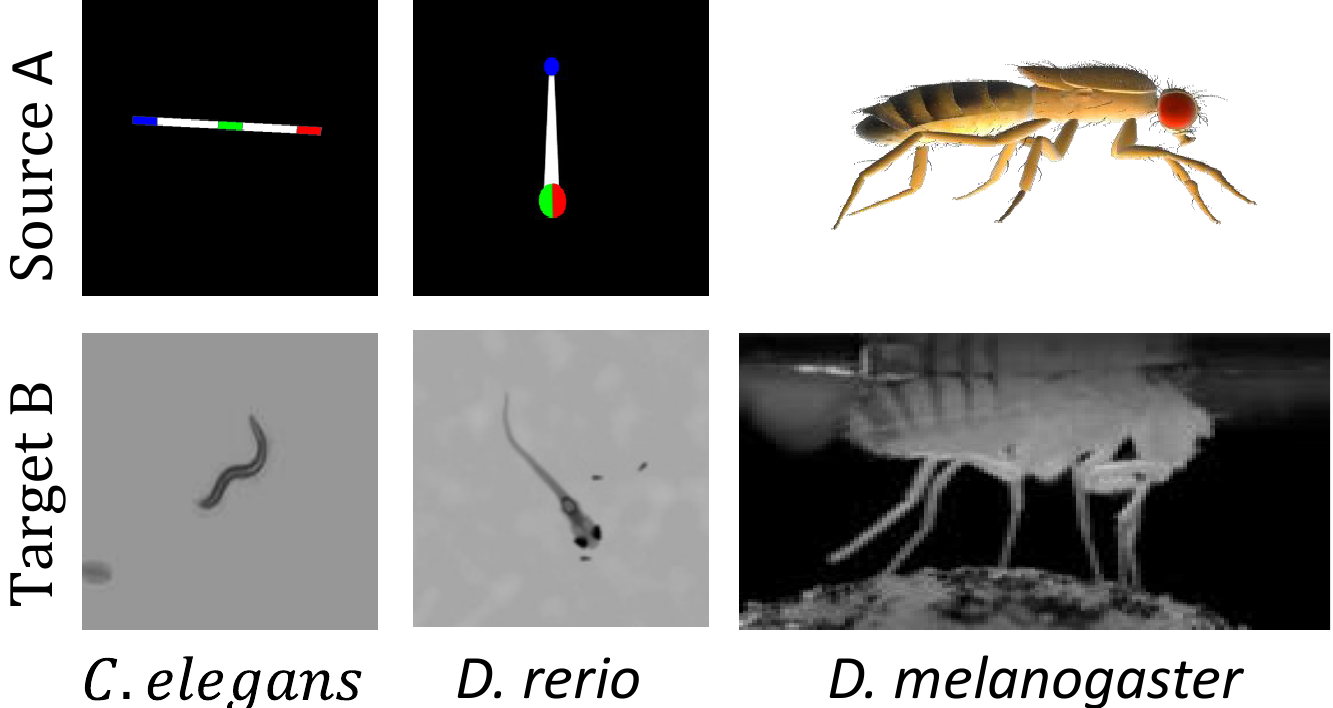}
	\caption{\textbf{Domain examples with large discrepancy in appearance, shape and pose.} Translating from rendering to real images requires bridging the domain gap without having pixel nor pose correspondences. It is particularly challenging in our setting, as even the realistic fly character shows significant differences in shape (body and limb width) as well as pose (legs stretched).}
	\label{fig:unpaired}
\end{figure}

In this paper, we introduce a novel approach to generate realistic images of different kinds of laboratory animals---flies, fish, and worms--from synthetic renderings for which labels such as keypoint annotations are readily available. The generated realistic images can then be used to train a deep network that operates on real images, as shown in Fig.~\ref{fig:teaser}. The challenge is to condition the generated images in such a way that the keypoints (e.g. skeleton joint positions) in the simulated source transfer to the realistic target; despite large differences in shape and pose as well as for small training sets that are practical, see Fig.~\ref{fig:unpaired}.

\begin{figure*}[t]
	\begin{center}
		\includegraphics[trim=1.1cm 1.8cm 0.1cm 0.1cm,clip,width=0.8\linewidth]{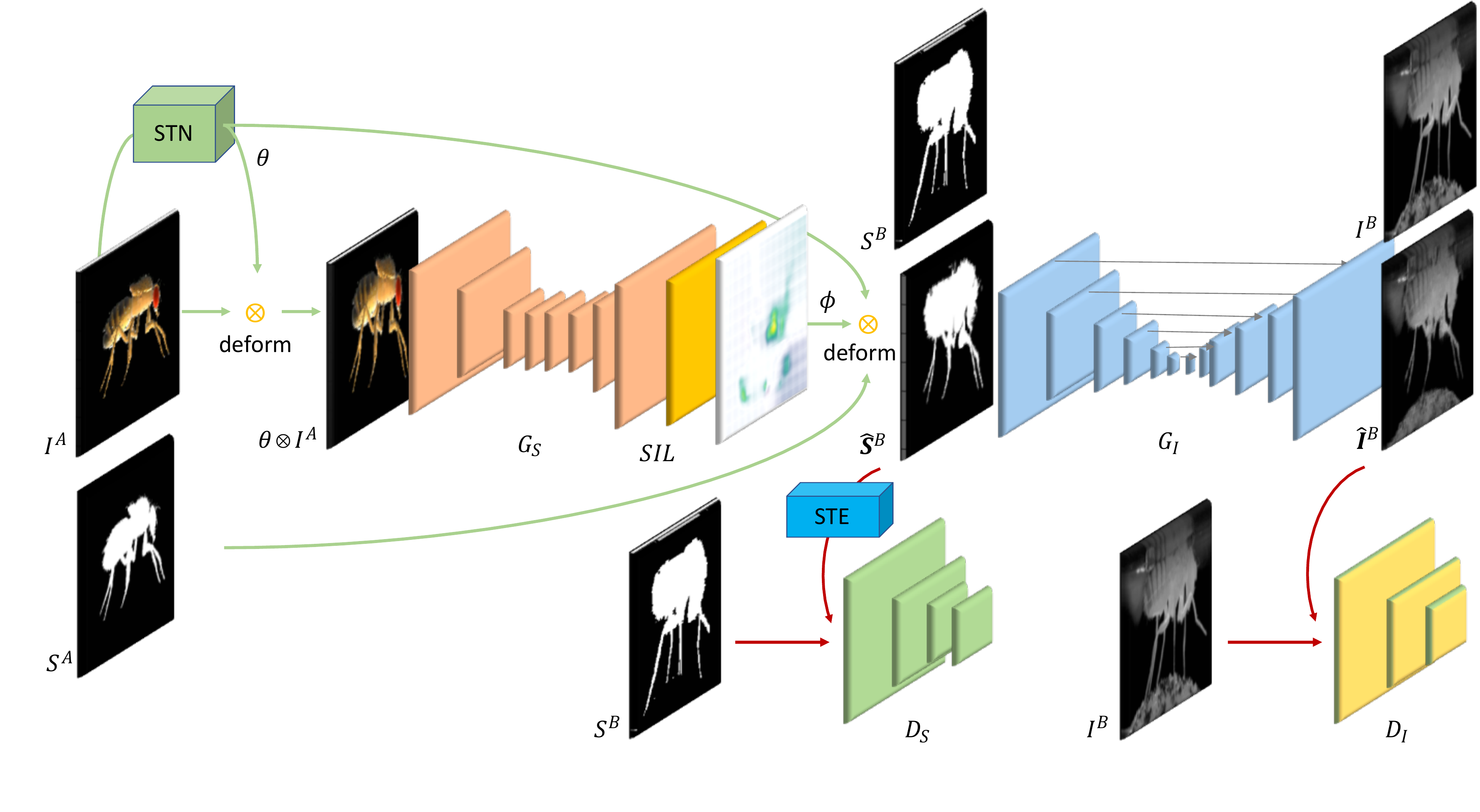}
	\end{center}
	\caption{\textbf{Overview of our deformation-based image translation method.} Our model has two steps. In the first step, the deformation from source domain A to target domain B is estimated for input image $\mI^A$ and it's silhouette $\mS^A$ via two networks $G_D$ and STN (Spatial Transformer Network). Their output is an explicit deformation field parameterized by the global, affine transformation $\theta$ and a local, non-linear warping $\phi$. Then, the deformed silhouette is transformed into the full output image $\hat{\mI}^B$ with image generator $G_I$. Discriminators $D_S$ and $D_I$ enable unpaired training. $D_S$ uses the Straight Through Estimator (STE) to backpropagate gradients through thresholding operations. }
	\label{fig:overview}
\end{figure*}

We model the change of 2D pose and shape in terms of a deformation field. This field is then paired with an image-to-image translator that synthesizes appearance while preserving geometry, as shown in Fig.~\ref{fig:overview}. 
Our approach is inspired by earlier approaches modeling human faces~\cite{Shu18} and brain scans~\cite{dalca2019learning}. We go beyond these studies in two important ways. First, we introduce silhouettes as an intermediate representation that facilitates independent constraints (loss terms) on shape and appearance.
It stabilizes training to succeed without reference images and helps to separate explicit geometric deformation from appearance changes.  
Furthermore, end-to-end training on unpaired examples is enabled with two discriminators and a straight-through estimator for non-differentiable thresholding operation and patch-wise processing. Second, to cope with large-scale as well as small-scale shape discrepancies, we introduce a hierarchical deformation model to separate global scaling, translation, and rotation from local deformation.

We test our method on flies (\textit{Drosophila melanogaster}), worms (\textit{Caenorhabditis elegans}) and  larval  zebrafish (\textit{Danio rerio}), see Fig.~\ref{fig:unpaired}, and compare it against state-of-the-art approaches that rely either on circularity constraint or hand-defined factorizations of style and content. Not only does our method generate more realistic images, but more importantly, when we use the images it generates to train pose estimators we get more accurate results. Nothing in our approach is specific to the animals we worked with and that could also be applied just as well to limbed vertebrates, including rodents and primates. 

\section{Related Work}

We present a method for spatially consistent image domain adaptation and pose estimation. In the following sections, we discuss recent advances towards this goal.

\parag{Pose Estimation.}
Deep learning based human pose estimation methods have recently made great progress. This is especially true for capturing human movements for which there is enough annotated data to train deep networks~\cite{Varol17,Ionescu14a,Lin14, Sigal10, Mahmood19,andriluka14}. A large corpus of the literature focuses on prediction of 2D key points from images directly~\cite{Newell16, Xiao18, Tang18, Insafutdinov16, Yang17, Xiu18}. There is also a wide literature on capturing 3D pose directly from images, or as a function of 2D keypoints instead~\cite{Popa17,Mehta17b,Rogez17,Pavlakos17,Zhou17d,Sun17,Pavllo19}. Weakly \cite{zhou2017towards} and semi-supervised algorithms \cite{Wandt_2019_CVPR}  can further improve the performance of motion capture systems, for example by using multi-view constraints~\cite{Rhodin19a,yao2019monet}.

Approaches designed primarily for human pose have been recently been transferred to study large animals, like cheetahs and lab mice~\cite{Nath19}. \cite{Zuffi19} uses a model based algorithm, trains on synthetic renderings, and refines on real zebra photographs. However, their quadruped body model does not translate to animals with a different number of legs and the suggested direct training on synthetic images for initialization did not succeed in our experiments, likely because realistic models are not available for our cases.

For pose estimation in \textit{Drosophila}, DeepLabCut provides a user-friendly interface to DeeperCut~\cite{Nath19}, LEAP~\cite{Pereira18} tracks limb and appendage landmarks, and DeepFly3D leverages multiple views to capture 3D pose \cite{Gunel19}. Nevertheless, all these methods require large amounts of manual labels, which are not available for many animals and cannot be reused when recording the same species in different environments and illumination conditions.

\parag{Paired Image-to-Image Translation.}
Supervised image-to-image translation methods aim to translate images across domains (e.g., day-to-night, summer-to-winter, photo-to-painting), often using adversarial methods \cite{isola2017image} to learn a mapping from input to output images. More recent studies have aimed to translate edges to images~\cite{sangkloy2017scribbler} and cascaded networks are used to condition on semantic label maps~\cite{chen2017photographic}. However, in our setting, no paired examples are available.

\parag{Style Transfer.}
Style transfer is an image-to-image translation method that works on unpaired examples, aiming to transfer the input image style while preserving the geometry of the target image~\cite{gatys2016image,johnson2016perceptual,gatys2016preserving,ulyanov2016instance}. Initial deep learning approaches optimized an image with respect to the Gram matrix statistics of deep features of the target image \cite{gatys2015neural,gatys2016image}. More recent studies have tested different architectures and loss functions \cite{li2016combining} and uses a contextual loss to transfer the style at the semantic level~\cite{mechrez2018contextual, Jing17}.

Our work is different from style transfer as we explicitly permit significant changes in pose and shape.

\parag{Unsupervised Domain Adaptation.}
Another line of work trains neural networks on unpaired examples for domain translation, including sim2real mappings. Early approaches used weight-sharing~\cite{liu2016coupled, zhu2017unpaired} and sharing of specific content features~\cite{bousmalis2017unsupervised,shrivastava2017learning}. The cycle consistency in Cycle-GAN, which assumes bijective mapping between two domains, can map from zebra to horse~\cite{zhu2017unpaired, kim2017learning,he2016dual}, but bridging large deformations across domains, such as for going from cat to dog and even more in our case (see Fig.~\ref{fig:unpaired}), requires alternative network architectures \cite{Gokaslan18} or intermediate keypoint representations \cite{Wayne19}; However,
none of the methods discussed above establish a fine-grained, dense spatial correspondence between source and target, which prevents accurate transfer of desired keypoint locations. 

\parag{Deformation networks.}
Explicit deformation has been used in diverse contexts. The spatial transformer network (STN) made affine and non-parametric spatial deformations popular as a differentiable network layer \cite{Jaderberg15}. These approaches have been used to zoom in on salient objects \cite{Recasens18}, disentangle shape and appearance variations in an image collection \cite{Shu18}, and register (brain scan) images to a common, learned template image \cite{dalca2019learning,balakrishnan19,kim2019unsupervised,rueckert2006diffeomorphic}. \cite{Fu2019} introduced global transformation into the Cycle-GAN framework.
While similar in spirit, additional advances beyond these approaches are still required to model deformations faithfully on our unpaired translation task.


\section{Method}

Our goal is to translate pose annotations and images from a synthetic domain $\dA$ to a target domain $\dB$ 
for which only unpaired images $\{\mI^\dA_i\}^N_{i=1}$ and $\{\mI^\dB_i\}^K_{i=1}$ exist. In our application scenario, the target examples are frames of video recordings of a living animal and the source domain are simple drawings or computer graphics renderings of a character animated with random deformations of the limbs. Both domains depict images of the same species, but in different pose, shape, and appearance.

Fig.~\ref{fig:overview} summarizes our approach. To tackle the problem of translating between domains while preserving pose correspondence, we separately transfer spatially varying shape changes via explicit deformation of the source images via an intermediate silhouette representation $\hat{\mS}^B$ (Section \ref{sec:space}). Subsequently, we locally map from silhouette to real appearance (Section \ref{sec:appearance}). The final goal is to train a pose estimator on realistic images from synthetic examples (Section \ref{sec:pose}). Our challenge then becomes to train neural networks for each, without requiring any paired examples or keypoint annotation on the target domain. 
To this end, we set up adversarial networks that discriminate differences with respect to the target domain statistics.
Learning of the image translation is performed jointly on the objective

\begin{equation}
\cL = \cL_I + \cL_S + \mathcal{R_D},
\end{equation}
where $\cL_I$ and $\cL_S$ are the adversarial losses on generated segmentation and image, and $\cR_D$ is a regularizer on the deformation grid. Besides images $\mI$, our method operates on segmentation masks $\mS$ of the same resolution. The domain origin is denoted with superscripts---$\mI^A$, and the domain target (real images) is denoted $\mI^B$. We use several generator and discriminator networks, which we denote $G$ and $D$, respectively, with subscripts differentiating the type---$G_I$. We explain each step in the following section.

\subsection{Spatial Deformation}
\label{sec:space}

\begin{figure}[t]
	\begin{center}
		\includegraphics[trim=0cm 0cm 0cm 0cm,clip,width=0.9\linewidth]{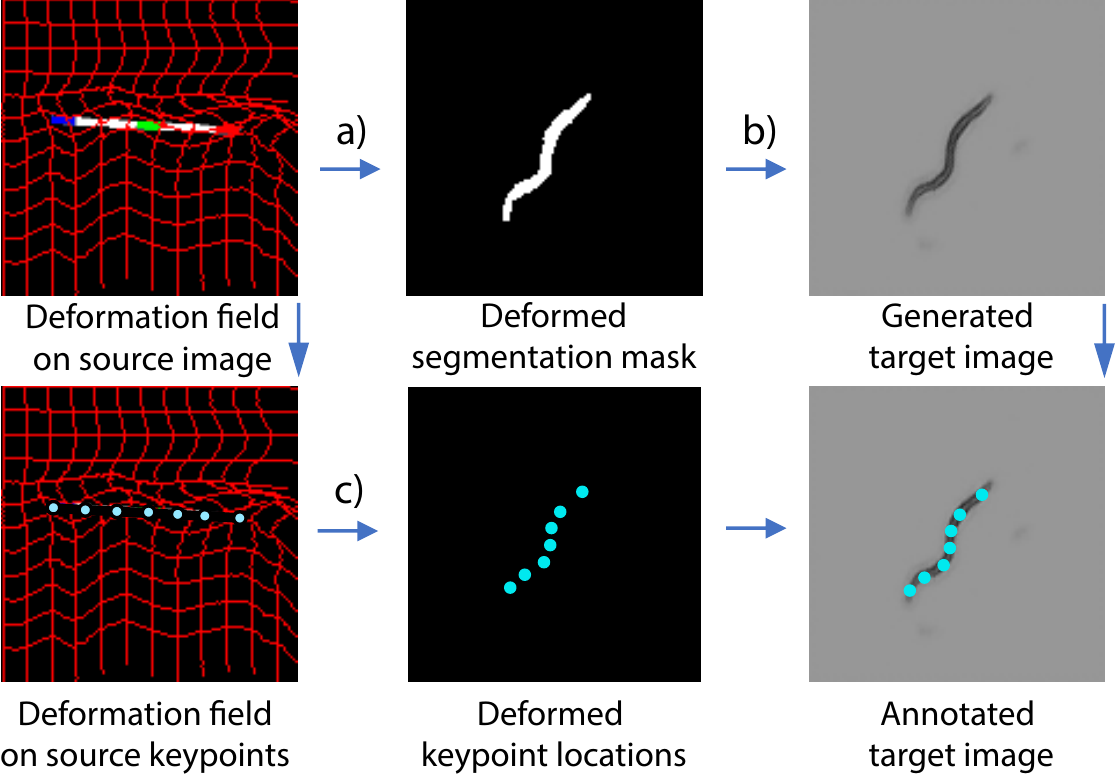}
	\end{center}
	\caption{\textbf{Explicit deformation ensures transfer of keypoints.} The deformation field is inferred as part of a) source image segmentation to target image segmentation transfer (including global, affine transformation) and b) segmentation to target image translation. c) The same deformation field is applied to transfer known keypoints from source to target. }
	\label{fig:deformation}
\end{figure}

Our experiments showed that using a single, large discriminator, as done by existing techniques, leads to overfitting and forces the generator to hallucinate, due to the limited and unrealistic pose variability of the simulated source.
We model shape explicitly through the intermediate silhouette representation and its changes with a per-pixel deformation field, as shown in Fig.~\ref{fig:deformation}.
The silhouette lets us setup independent discriminators with varying receptive field; large for capturing global shape and small to fill-in texture. Moreover, the deformation field enables the desired pose transfer while bridging large shape discrepancies.

The first stage is a generator $G_S$ that takes a synthetic image $\mI^A$ and mask ${\mS}^A$ as input, and outputs a deformed segmentation mask $\hat{\mS}^B$ that is similar to the shapes in $B$. 
To model global deformation, we use a spatial transformer network (STN)  \cite{Jaderberg15} that takes the synthetic image $\mI^\dA  \in \R^{C,H,W}$ as input, and outputs an affine matrix $\theta \in \R^{3,4}$, which models global scaling, translation and rotation differences between the source and target domains. It is trained jointly with a fully-convolutional generator network, $G_D$, which takes the globally transformed image as input and outputs $\phi \in \R^{2,H,W}$, a per-pixel vector field that models fine-grained differences in pose and shape. 
The vector at pixel location $x$ in $\phi$ points to the pixel in the source domain that corresponds to $x$. Overlaying the source pixels of selected rows and columns of $\phi$ leads to the deformed grid visualized in Fig.~\ref{fig:deformation}.
This hierarchical representation allows us to cope with varying degrees of discrepancies between the two domains. We refer to the combined application of these two networks and their output as generator $G_S(\mI^A,\mS^A) = \phi \otimes \theta  \otimes \mS^A$, where $\theta = STN(\mI^A)$, $\phi = G_D(\theta \otimes \mI^A)$, and $\otimes$ denotes the transformation by global and local deformation.

Training $G_S$ requires silhouettes in $\dA$ and $\dB$. Silhouettes $\mS^A$ in the source domain are trivially obtainable from synthetic characters by rendering them on a black background. It is relatively easy to estimate $\mS^B$ on a static background for the target domain as datasets are obtained in controlled lab environments. We will later demonstrate that our model is robust to remaining errors in segmentation. 

The difficulty of our task is that all domain examples are unpaired, hence, a constraint can only be set up in the distributional sense. Thus, we train a shape discriminator $D_S$ alongside $G_S$ and train them alternately to minimize and maximize the adversarial loss
\begin{align}
&\cL_S = \cL_{GAN}(G_{\mS},D_{\mS},\mS^\dA,\mS^\dB) \\
&=\mathbb{E}_{\mS^B}[\log D_S(\mS^B)]
 +\mathbb{E}_{\mS^A}[\log(1-D_S(G_S(\mS^A)))],\nonumber
\end{align}  
where the expectation is built across the training set of $A$ and $B$. The adversarial loss is paired with the regularizer 
\begin{equation}
\mathcal{R}_D =  \alpha(\left \| \bigtriangledown \phi_x(A) \right \|^{2} + \left \| \bigtriangledown \phi_y(A)) \right \|^{2}) + \beta \left \| \phi(A) \right \|,
\end{equation}
to encourage smoothness by penalizing deformation magnitude and the gradients of the deformation field, as in \cite{Shu18}.

The inputs of the discriminator are binary masks from source domain $\dA$ and target domain $B$. However, the deformed masks are no longer binary on the boundary because of the interpolation required for differentiation. Thus, it would be trivial for $D_S$ to discriminate against the real and synthesized masks based on non-binary values. To overcome this issue, we threshold to get a binary mask. Although the threshold operation is not differentiable, we can still estimate the gradients with respect to $G_S$ using a straight through estimator (STE) \cite{yin2018understanding}, which treats the threshold as the identity function during backpropagation, and therefore passes the gradients on to the previous layer.

\parag{Implementation details.} Directly outputting a vector field leads to foldovers that make the training unstable. Instead, we parameterize it as the gradient of the deformation field $\phi$, and enforce positivity to prevent foldovers as in \cite{Shu18}. $\phi$ can be recovered by summing the gradients across the image. The deformation from $\dA$ to $\dB$ is implemented with a spatial transformer layer (STL) that infers the value of deformed pixel locations by bilinear interpolation \cite{Jaderberg15} and is differentiable. In contrast to \cite{Shu18}, we use a fully convolutional network to learn the local deformation field. The $G_D$ network consists of 3 Resnet blocks between downsampling/upsampling layers. The receptive field of the network is 64 pixels, $1/2$ of the image, which is sufficient for our experiments.

The STN network consists of 5 convolutional layers and a fully connected stub to output $\theta$ that is preceded by max-pooling and SELU units (this yielded better results in preliminary experiments, compared to ReLU activations).

\subsection{Appearance Transfer}
\label{sec:appearance}

Once the shape discrepancies between the two domains have been estimated and corrected by $G_S$, we then generate the appearance of the target domain on the deformed silhouettes $\hat{\mS}^B = G_S(\mI^A,\mS^A)$.
We deploy a generator $G_I$ that is configured to preserve the source shape, only filling in texture details. The input is $\hat{\mS}^B$ and the output is a realistic image $\hat{\mI}^B$ that matches the appearance of the target domain. We use a discriminator $D_I$ for training, as synthetic and real images are unpaired. In addition, our choice of using the silhouette as an intermediate representation allows us to introduce a supervised loss on $\mS(\mI^B)$ computed from real images $\mI^B$. The training objective is
\begin{align}
\cL_I & = \cL_{GAN}(G_I,D_I,\mI^\dA,\mI^\dB) + \left \| G_I(\mS(\mI^B)) - \mI^B \right \|,
\end{align}
where the GAN loss is defined as before and the second part is the supervised loss which stabilizes training.

\begin{figure}[t]
	\begin{center}
		\includegraphics[trim=0cm 0cm 0cm 0cm,clip,width=0.8\linewidth]{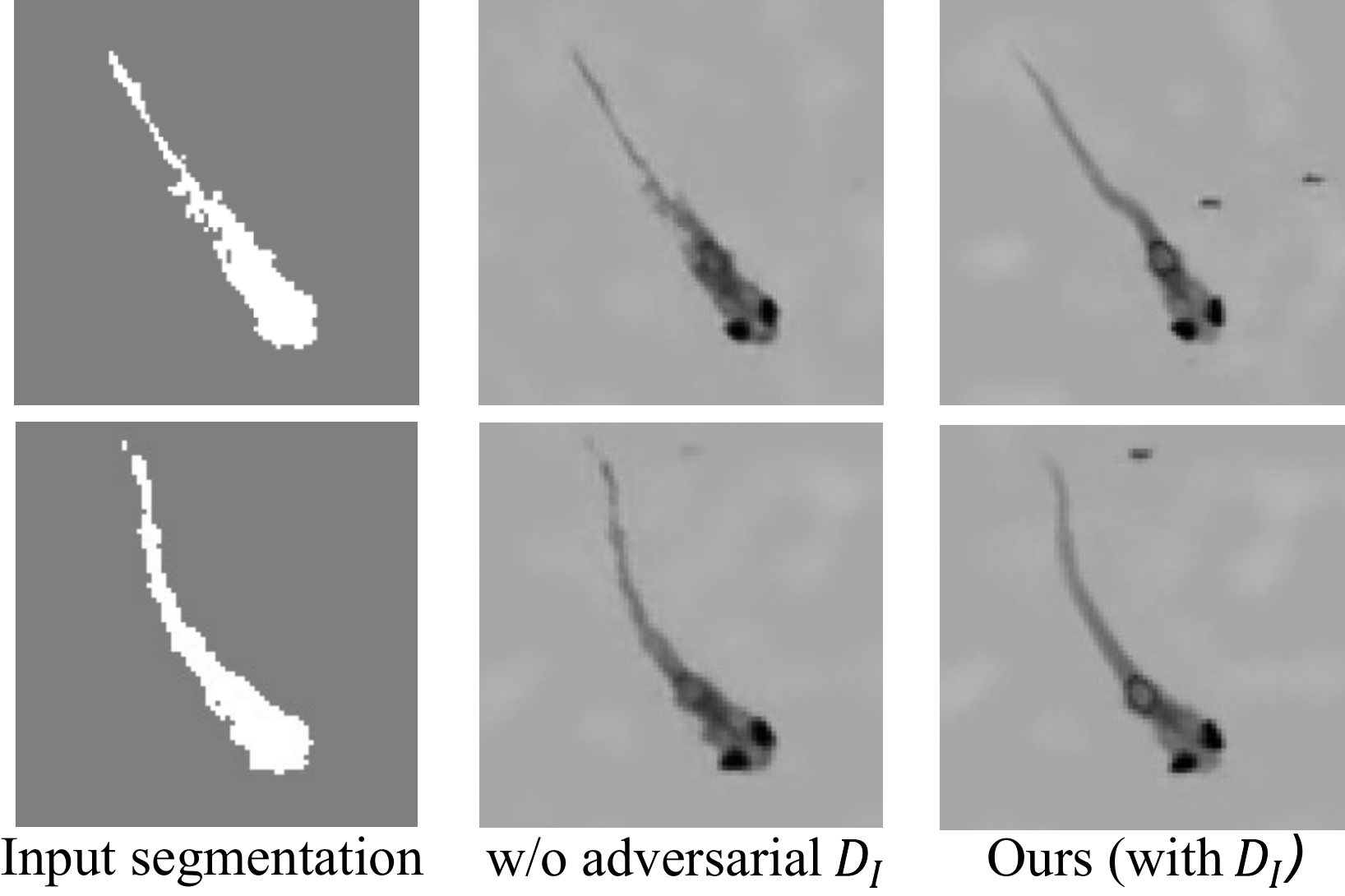}
	\end{center}
	\caption{\textbf{Texture discriminator influence.} Without the adversarial discriminator, the image generator is disturbed by an irregular silhouette boundary. In our model, the adversarial $D_I$ creates a link to the deformed silhouettes $\hat{S}^B$ enabling end-to-end training.}
	\label{fig:ablation wo DI}
\end{figure}

Training the supervised loss in isolation without end-to-end training with the adversarial losses leads to artifacts since neither the synthesized nor silhouettes from real images are perfect, see Fig.~\ref{fig:ablation wo DI}.

The pose distribution of the simulated character can differ even after local and global deformation as some pose differences cannot be explained by an image deformation. For instance, the occlusion effects of crossing legs on \textit{Drosophila} cannot be undone as a 2D image transformation. A discriminator with a large receptive field could detect these differences and re-position legs at locations without correspondence in the source. To counteract this issue, we make sure $D_I$ has a small receptive field. This is possible without suffering from texture artifacts since the global shape deformation is already compensated by $G_S$ and the texture can be filled in locally.

\parag{Implementation details.} 
We use a 7-layer U-Net generator as our backbone network for image translation with $G_I$. The skip connections in the U-Net help the network  preserve the spatial information. For $D_I$, we use a patch-wise discriminator, consisting of three 4x4 convolutional layers; the first one with stride two and the second one with instance normalization. All activation functions are leaky ReLU.
The small receptive field of the patch discriminator additionally helps to maintain the spatial structure of the object and was sufficient in our experiment to reproduce the real appearances faithfully. 

\subsection{Pose Estimation}
\label{sec:pose}

We use the stacked hourglass network architecture for pose estimation~\cite{Newell16}. Stacked hourglass is a fully-convolutional network with several bottlenecks that takes an image $\mI$ and outputs a heatmap $\mH$ of the same aspect ratio but at four times lower resolution due to pooling at the initial layers. The heatmaps $\mH$ are a stack of 2D probability maps with Gaussian distribution, where the maximum value of each channel in the stack indicates one specific joint location.
Because our source images are synthesized from 3D character models, we can use the virtual camera matrix to project 3D keypoints, such as the knee joint, onto the image.

To obtain annotations in the target domain, we conveniently use the image deformation operation $\mH_d = (\phi, \theta) \otimes \mH$ to compute the deformed heatmap $\mH_d$ that matches to the synthesized target domain image $\mI_d = G_I(G_S(\mI^A,\mS^A) = G_I((\phi, \theta) \otimes \mI^A)$, with $\phi$ coming from $G_D$ and $\theta$ from the STN in $G_S$. Note, that this is only possible due to the explicit handling of deformations.

Having synthesized realistic examples of the target domain and transferred ground truth heatmaps, it remains to train the pose estimation network in a supervised manner. We use the $L_2$ loss between the predicted and ground truth heatmaps.
At test time, we estimate the corresponding joint location as the argmax of the predicted heatmap, as usual in the pose estimation literature. Because the worm is tail-head symmetric, we compute errors for front-to-back and back-to-front ordering of joints and return the minimum at training and test time. We call this a permutation invariant (PI) training and testing. In the same vein, we regard the correct assignment of the six \textit{Drosophila} legs as an independent task that is extremely hard to solve in the 2D domain. To separate and sidestep this problem, we compute the test error for all possible permutations and return the minimum.

\parag{Implementation details.} Input images are augmented by random rotations, drawn uniformly from $[-30^\circ, 30^\circ]$.
Additional details are given in the supplemental document.


\section{Evaluation}

In this section, we qualitatively compare our results to canonical baselines and variants of our algorithm, in order to highlight advantages and remaining shortcomings both visually and quantitatively.  This includes the task of 2D keypoint localization on the target domain.
We test our approach on different neuroscience model organisms in order to demonstrate varying complexity levels of deformation and generality to different conditions. Additional qualitative results and comparisons are given in the supplemental document.

All input and output images are of dimension $(128,128)$.
We operate on gray-scale images, i.e. channel dimension $C=1$, obtained from infrared cameras, which are commonly used in neuroscience experiments in order to avoid inadvertent visual stimulation. Nevertheless, our method extends naturally to color images.

\begin{figure}[t]
	\begin{center}
		\includegraphics[trim=0cm 24.1cm 20.cm 0cm,clip,width=1.0\linewidth]{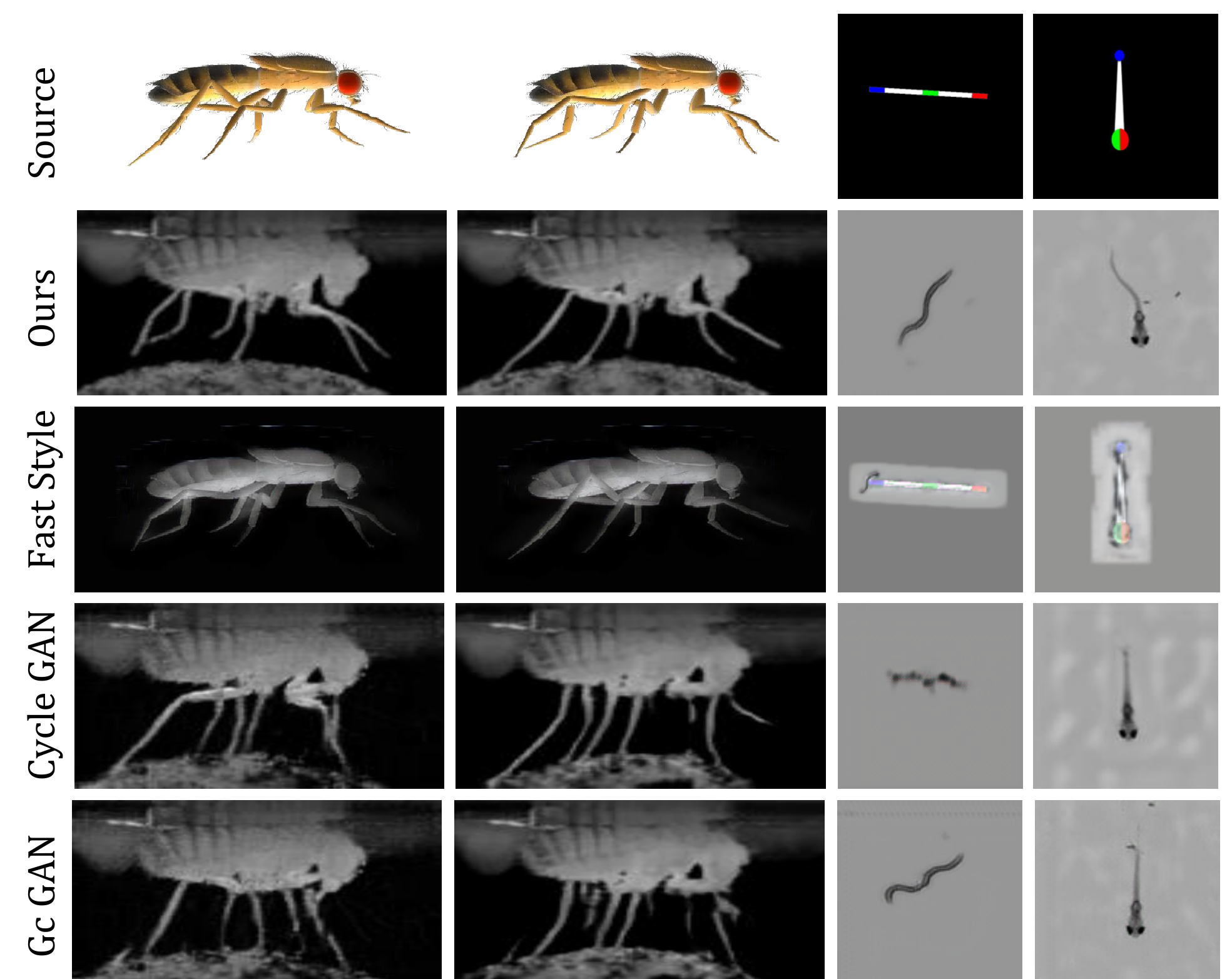}
	\end{center}
	\caption{\textbf{Qualitative comparison.} Existing unpaired image translation methods can generate realistic images on worm and fish, but exhibit artifacts for the thin legs of the \textit{Drosophila} and zebrafish examples. Ours succeeds on all three classes.}
	\label{fig:quality}
\end{figure}

\parag{Datasets.}
We test on available zebrafish and worm image datasets, by \cite{Johnson19} and \cite{Yemini13,Balazs14}, using 500 and 100 real images for unpaired training. To quantify pose estimation accuracy, we manually annotate a test set of 200 frames with three keypoints (tail and eyes) for the zebrafish and two points (head and tail) for the worm. In these datasets, the background is monochrome and is removed by color keying to obtain the foreground masks. Because of the simplicity of these models, we use a simple, static stick figure as a source image that is augmented by uniformly random rotation and translation. Fig.~\ref{fig:quality} gives example images.

Our most challenging test case is the \textit{Drosophila} fly. We use the subset of the dataset published alongside \cite{Gunel19}, which
contains transitions between different speeds of walking, grooming and standing captured from a side view and includes
 annotations for five keypoints for each of the fully-visible legs (four joints and tarsus tip).
In this dataset, the fly is tethered to a metal stage of a microscope and the body remains stationary, yet the fly can walk on a freely rotating ball (spherical treadmill), see Fig.~\ref{fig:quality}. To get the target domain segmentation masks, we first crop out the ball and background clutter with a single coarse segmentation mask. This mask is applied to all images due to the static camera setup. The body, including the legs, is then segmented by color keying on the remaining black background. Please note, that at test time, no manual segmentation is used.

We use 815 real images for unpaired training and 200 manually annotated images for testing.
On the source side, we render 1500 synthetic images using an off-the-shelf Maya model from \href{https://www.turbosquid.com/3d-models/drosophila-melanogaster-fruit-fly-obj/576944}{turbosquid.com}. The source motion is a single robotic walk cycle from \cite{ramdya2017climbing} which we augment by adding random Gaussian noise to the character control handles. This increases diversity but may lead to unrealistic poses that our deformation network helps to correct.

\parag{Metrics.} 
The pose estimation accuracy is estimated as the root mean squared error (RSME) of predicted and ground truth 2D location and percentage of correct keypoints (PCK), the ratio of predicted keypoints below a set threshold. We report results for thresholds ranging from 2 to 45 pixels. We also provide accumulated error histograms and the average PCK difference as the area under the curve (AUC) of the error histogram, to analyze the consistency of the improvements. 

In many cases, it is impossible, even for a human, to uniquely identify the leg identity for \textit{Drosophila}. As in \cite{Gunel19}, we therefore only evaluate the three entirely visible legs. Moreover, we find the optimal leg assignment across the three legs at test time as PI-RSME, PI-PCK, and PI-AUC, using the permutation invariant metric introduced in Section~\ref{sec:pose}.
Because the images of the worm are front-back symmetric, we train and test by permuting keypoints front-to-back and back-to-front. 
The pose estimation task lets us quantify the made improvements, both due to more realistically generated images (image quality), as well as the preservation of correspondences (geometric accuracy) since the lack of one would already lead to poor pose estimation.

To independently quantify the image quality, we use the structural similarity (SSIM) index  \cite{Wang04imagequality}. We measure the similarity between all generated images $\hat{\mI}^B$ (for every $\mI^A$ in A) with a pseudo-randomly sampled reference image $\mI^B$.

\parag{Baselines.} 
We compare to Fast-Style-Transfer~\cite{Engstrom2016}, which combines \cite{gatys2015neural,johnson2016perceptual,ulyanov2016instance}, Cycle-GAN~\cite{zhu2017unpaired} and Gc-GAN~\cite{Fu2019}. With the latter being a state-of-the-art method for image to image translation and the former used to validate that simpler solutions do not succeed.
We compare pose estimation with the same architecture, trained directly on the synthetic images, images generated by the above mentioned methods, and on manual annotations of real training images (185 for Drosophila, 100 for worm, and 100 for fish).

\subsection{Quality of Unpaired Image Translation}
The quality of Cycle and Gc-GAN is comparable to ours on the simple worm and fish domains, as reflected visually in Fig.~\ref{fig:quality} and quantitatively in terms of SSIM in Table~\ref{tab:quality}. For \textit{Drosophila}, our method improves image quality (0.66 vs.\ 0.39, 0.63 and 0.65). Albeit the core of explicit deformation was to transfer pose annotations across domains, this analysis shows that an explicit mapping and incorporation of silhouettes regularizes and leads to improved results. For instance, it ensures that thin legs of the fly are completely reconstructed and that exactly six legs are synthesized, while Cycle-GAN and Gc-GAN hallucinate additional partial limbs.

\begin{table}[t]
\centering
\resizebox{0.7\columnwidth}{!}
{
	\begin{tabular}{|llll|}
		\hline
		\multicolumn{1}{|l|}{Task} & \it D.M. & \it C.E. & \it D.E. \\ \hline
		Fast-Style-Transfer & 0.3932 & 0.0539 & 0.6385 \\
		Cycle-GAN & 0.6543 & 0.9034 & 0.8504 \\
		Gc-GAN & 0.6392 & 0.8915 & 0.8586 \\
		Ours & \textbf{0.6619} & \textbf{0.9143} & \textbf{0.8847} \\ \hline
	\end{tabular}
}
\smallskip
\caption{\textbf{Structured similarity (SSIM) comparison}. The explicit modeling of deformation outperforms baselines, particularly on the complex \textit{Drosophila} images showing complex poses.
\label{tab:quality}
}
\end{table}

\subsection{Pose Domain Transformation}

\begin{figure}[t]
	\begin{center}
		\includegraphics[width=1\linewidth]{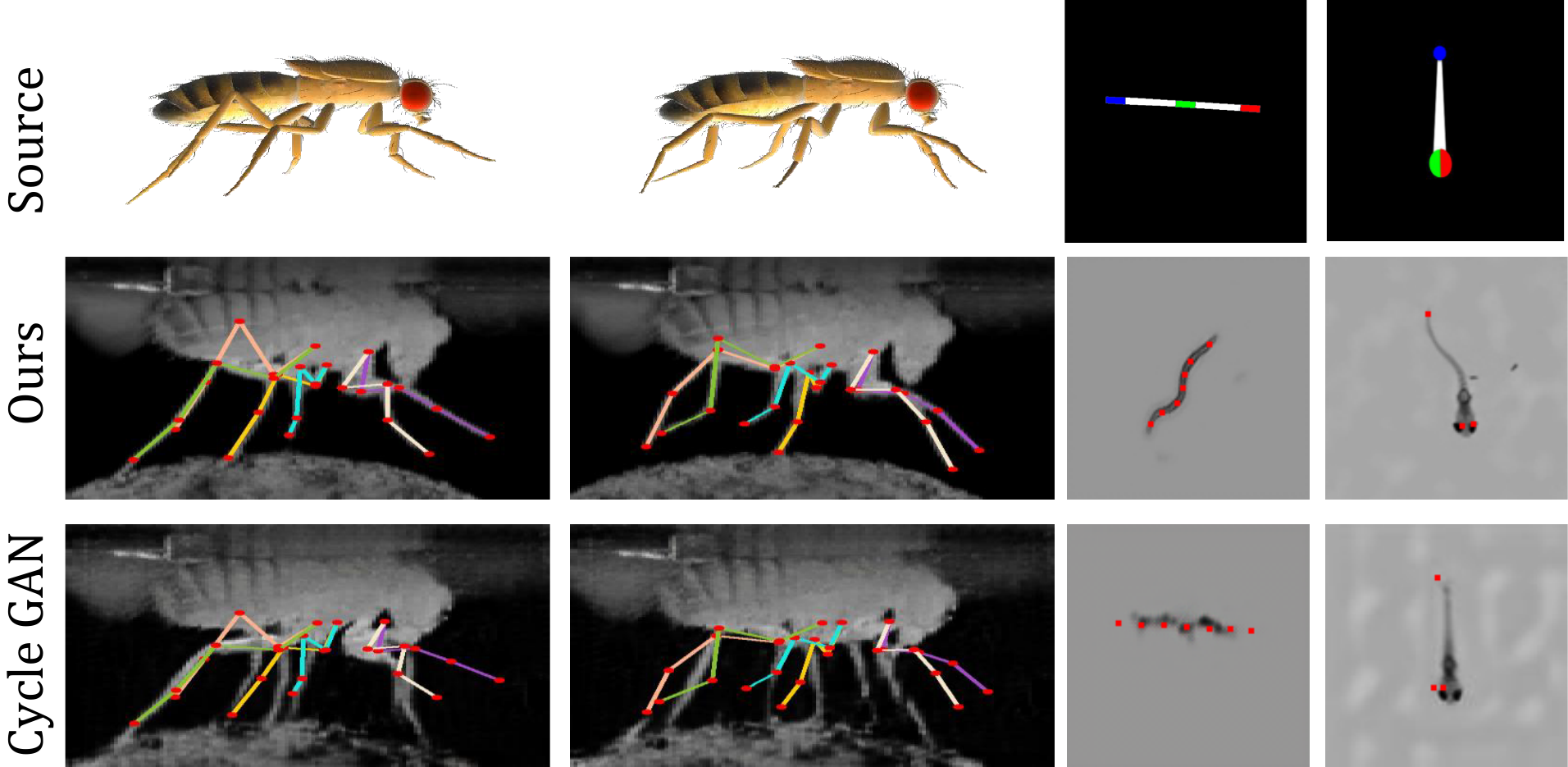}
	\end{center}
	\vspace{-8pt}
	\caption{\textbf{Automatic Pose Annotation.} Our method faithfully transfers poses across domains, while Cycle-GAN, the best performing baseline, loses correspondence on all three datasets.}
	\label{fig:pose}
\end{figure}

Fig.~\ref{fig:pose} shows that our method faithfully transfers 2D keypoints, obtained for free on synthetic characters, to the target domain. The transferred head and tail keypoints on the worm and fish correspond precisely to the respective locations in the synthesized images, despite having a different position and constellation in the source. This transfer works equally well for the more complex \textit{Drosophila} case. Only occasional failures happen, such as when a leg is behind or in front of the torso, rendering it invisible in the silhouette.
Moreover, the eyes of the fish are not well represented in the silhouette and therefore sometimes missed by our silhouette deformation approach. 

By contrast, existing solutions capture the shape shift between the two domains, but only implicitly, thereby loosing the correspondence. Poses that are transferred one-to-one from the source do no longer match with the keypoint location in the image. Keypoints are shifted outside of the body, see last column of Fig.~\ref{fig:pose}.
The style transfer maintains the pose of the source, however, an appearance domain mismatch remains. We show in the next section that all of the above artifacts lead to reduced accuracy on the downstream task of pose estimation.

\begin{figure}[t]
	\begin{center}
		\includegraphics[width=1\linewidth]{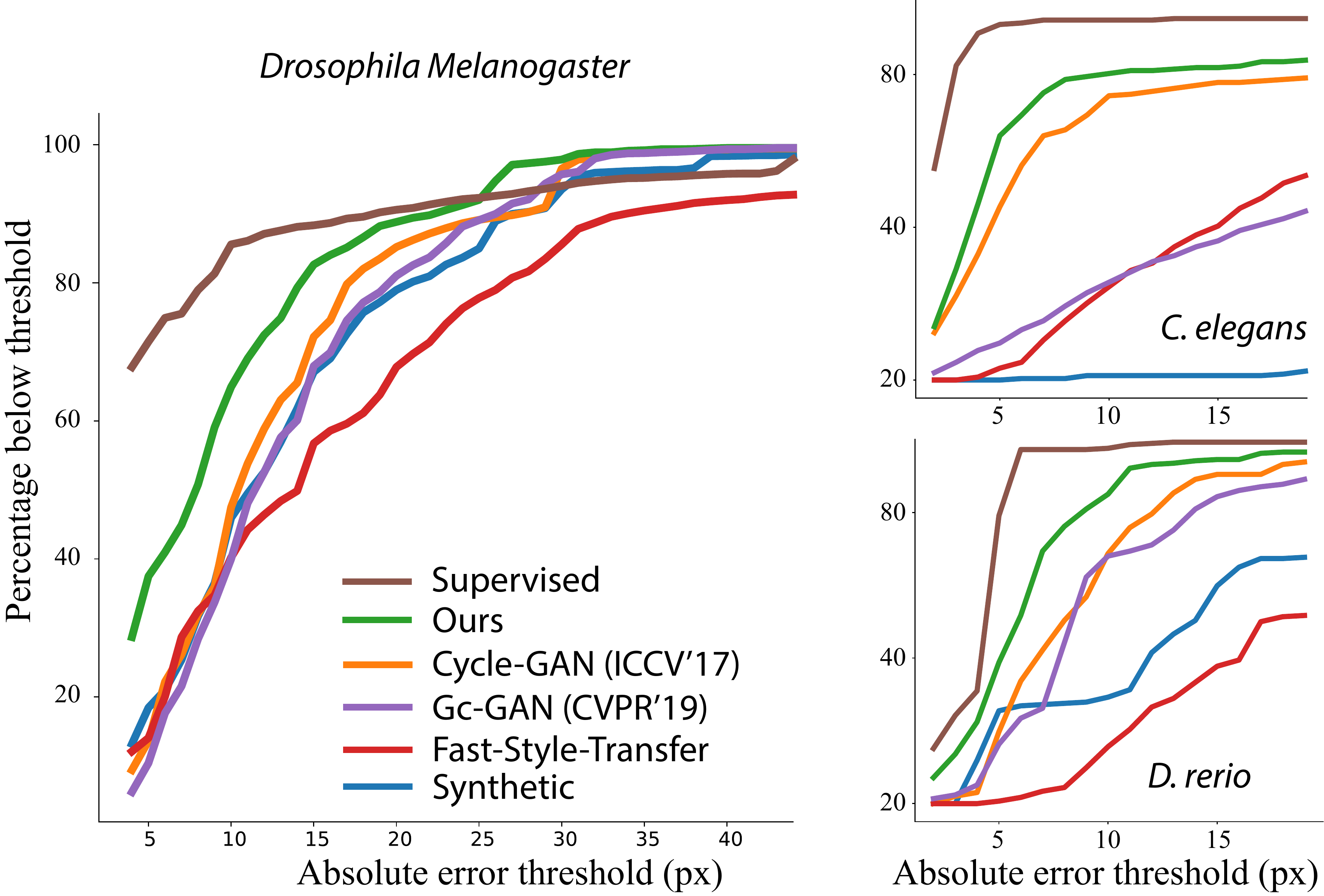}
	\end{center}
	\vspace{-8pt}
	\caption{\textbf{Pose estimation accuracy.} The accumulated error curves show the accuracy (vertical axis) for different PCK thresholds (horizontal axis). Our method clearly outperforms the baselines and approaches the manually supervised reference. }
	\label{fig:auc}
\end{figure}

\begin{table}[]
	\centering
	\resizebox{0.9\columnwidth}{!}
	{
		\begin{tabular}{|l|cccc|}
			\hline
			&\multicolumn{4}{|c|}{\it Drosophila Melanogaster} \\ \hline
			\multicolumn{1}{|l|}{Metric} & \multicolumn{1}{l|}{\begin{tabular}[c]{@{}l@{}}PI-PCK $\uparrow$\\ (5 pix)\end{tabular}} & \multicolumn{1}{l|}{\begin{tabular}[c]{@{}l@{}}PI-PCK $\uparrow$\\ (15 pix)\end{tabular}} & \multicolumn{1}{l|}{\begin{tabular}[c]{@{}l@{}}PI-AUC $\uparrow$\\ (4-45 pix)\end{tabular}} & \begin{tabular}[c]{@{}l@{}}PI-RMSE $\downarrow$\\ (pix)\end{tabular} \\ \hline
			Synthetic & 19.8 & 67.9 & 75.75 & 13.456 \\
			Fast-Style-Transfer & 15.4 & 57.6 & 68.9 & 17.309 \\
			Gc-GAN & 11.9 & 68.7 & 76.3 & 13.175 \\
			Cycle-GAN & 15.0 & 72.9 & 78.4 & 12.302 \\
			Ours & \textbf{38.6} & \textbf{83.2} & \textbf{85.1} & \textbf{9.289} \\
			\hline
			Supervised & 72.2 & 88.8 & 90.35& 6.507 \\
			\hline
		\end{tabular}
	}
	\smallskip
	\caption{\textbf{Pose estimation accuracy comparison on \textit{Drosophila Melanogaster}.} A similar improvement as for Drosophila is attained on the other tested laboratory animals, with a particularly big improvements on the zebrafish. Pose-invariant training improves results.}
	\label{tab:poseD}
\end{table}

\begin{table}[]
	\centering
	\resizebox{1\columnwidth}{!}
	{
		\begin{tabular}{|l|ccc|ccc|}
			\hline
			&\multicolumn{3}{|c|}{\it Caenorhabditis elegans} & \multicolumn{3}{|c|}{\it Danio rerio}\\ \hline
			\multicolumn{1}{|l|}{Metric} & \multicolumn{1}{l|}{\begin{tabular}[c]{@{}l@{}}PI-PCK $\uparrow$\\ (5 pix)\end{tabular}} & \multicolumn{1}{l|}{\begin{tabular}[c]{@{}l@{}}PI-AUC $\uparrow$\\ (2-20 pix)\end{tabular}} & \begin{tabular}[c]{@{}l@{}}PI-RMSE $\downarrow$\\ (pix)\end{tabular} & \multicolumn{1}{l|}{\begin{tabular}[c]{@{}l@{}}PCK $\uparrow$\\ (10 pix)\end{tabular}} & \multicolumn{1}{l|}{\begin{tabular}[c]{@{}l@{}}AUC $\uparrow$\\ (2-20 pix)\end{tabular}} & \begin{tabular}[c]{@{}l@{}}RMSE $\downarrow$\\ (pix)\end{tabular}\\ \hline
			Synthetic & 0.0 & 0.9 & 67.29 & 29.3 & 37.4 & 20.15 \\
			Fast-Style-Transfer & 3.1 & 25.0 & 20.50 & 15.6 & 20.8 & 19.25 \\
			Gc-GAN & 9.7 & 25.0 & 27.38 & 68.2 & 54.5 & 27.38 \\
			Cycle-GAN & 45.3 & 63.2 & 14.71 & 68.7 & 59.1 & 9.70 \\
			Ours & \textbf{64.0} & \textbf{70.9} & \textbf{11.17} & \textbf{85.0} & \textbf{72.1} & \textbf{7.23} \\
			\hline
			Supervised & 93.1 & 91.3 & 4.15 & 97.6 & 84.9 & 4.35\\
			\hline
		\end{tabular}
	}
	\smallskip
	\caption{\textbf{Pose estimation accuracy on \textit{C. elegans} and \textit{D. rerio}.} Our method significantly outperforms all baselines and approaches the supervised baseline. Units are given in round brackets.}
	\label{tab:poseWZ}
\end{table}

\subsection{2D Pose estimation}

\begin{figure}[t]
	\begin{center}
		\includegraphics[trim=0cm 0cm 0cm 0cm,clip,width=0.9\linewidth]{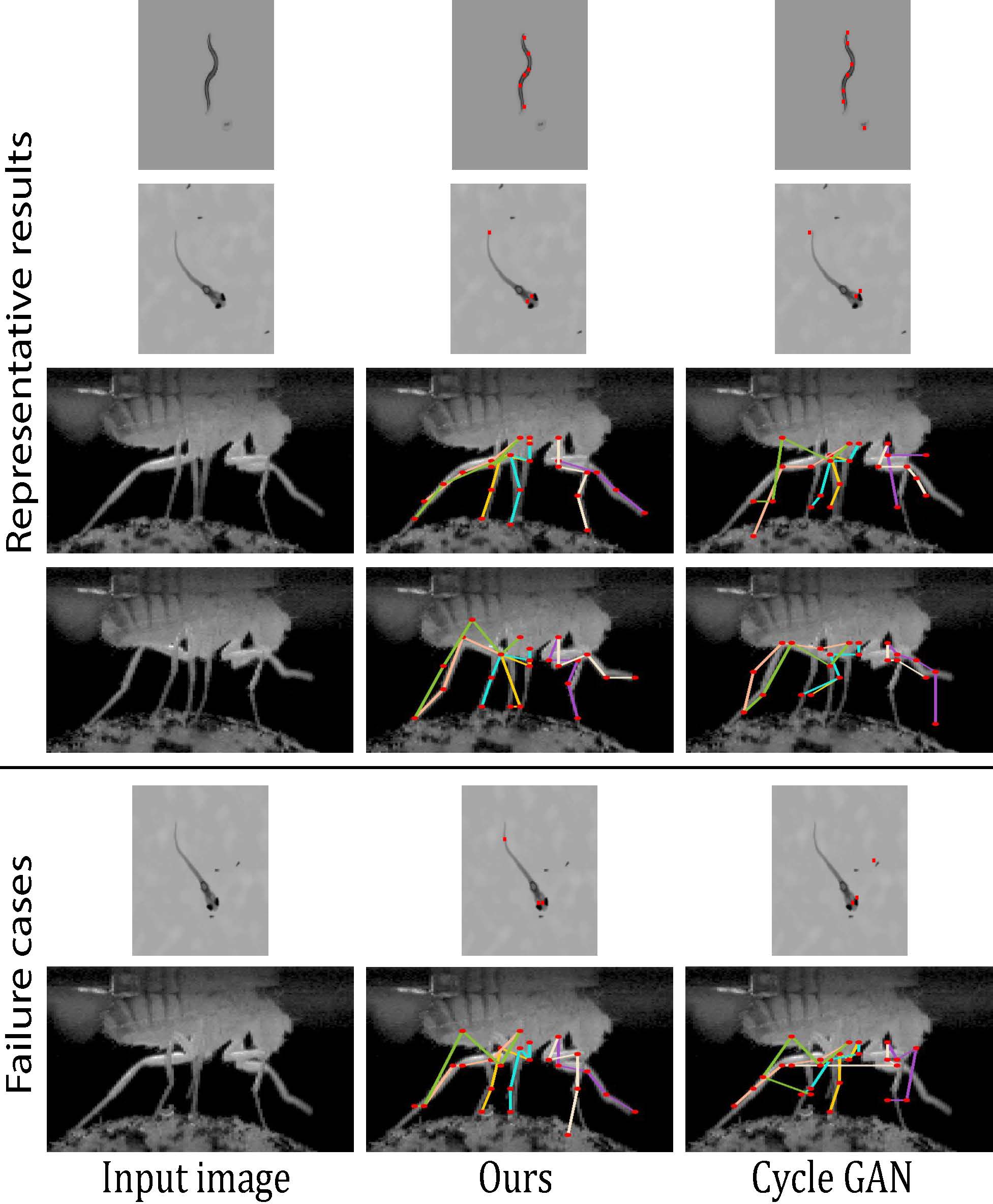}
	\end{center}
	\vspace{-8pt}
	\caption{\textbf{Qualitative pose estimation results.} The estimator provides decent results across all three animals. Occasional failures (last two rows) happen when legs cross, at occlusions, and for the fine fish tail. Training on Cycle-GAN images does not succeed.}
	\label{fig:qualitative pose}
\end{figure}

The primary objective of this study is to demonstrate accurate keypoint detection on a target domain for which only annotations on synthetic images with different shape and pose exist. 
Fig.~\ref{fig:qualitative pose} shows qualitative results. We compare the performance of the same keypoint detector trained on images and keypoints generated by ours and the baseline methods.
The absolute errors (tables~\ref{tab:poseD} and \ref{tab:poseWZ}) and accumulated error histograms (Fig.~\ref{fig:auc}) show significant (PCK 15: 83.2 vs. 72.9 Cycle-GAN)  and persistent (AUC 85.1 vs 78.4) improvements for Drosophila and the other domains. A similar improvement is visible for the simpler worm and zebrafish datasets, with even higher gains of up to 13 PCK points.
Although there remains a gap compared to training on real images with manual labels for small error thresholds, our method comes already close to the supervised reference method in PCK 15 and above. Compared to existing unpaired image translation methods, we gain a large margin.

The effect of the introduced explicit and hierarchical deformation model and the two-stage shape-separating training is analyzed independently in the following section.

\subsection{Ablation study}

\begin{figure}[t]
	\begin{center}
		\includegraphics[trim=0cm 0cm 0cm 0cm,clip,width=1\linewidth]{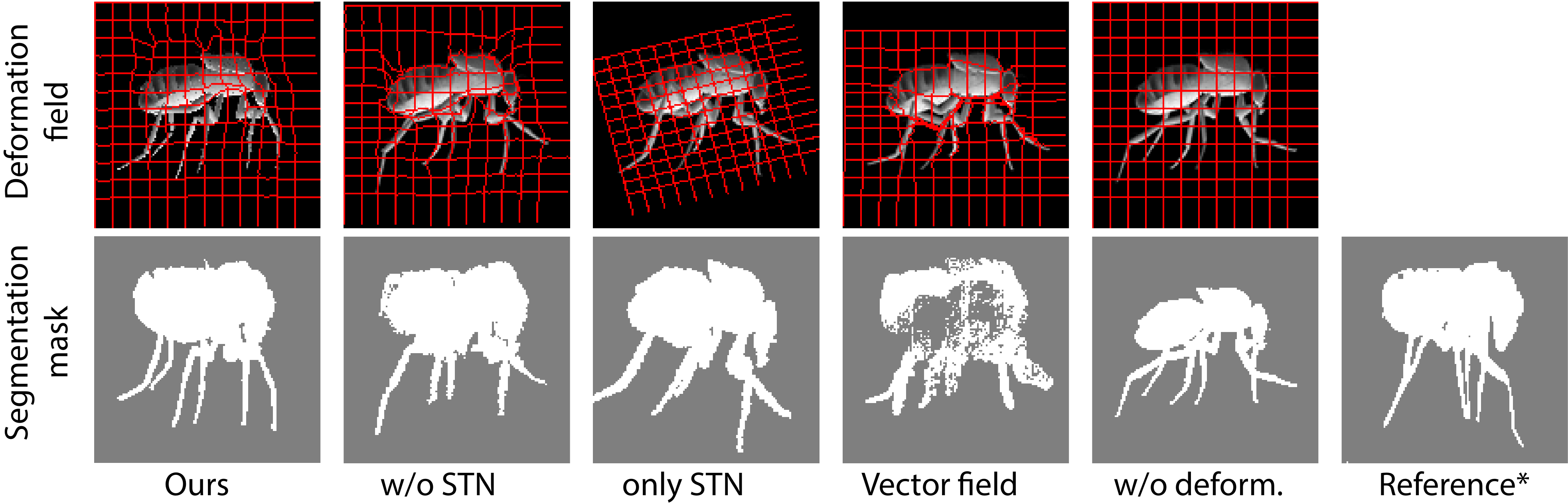}
	\end{center}
	\vspace{-8pt}
	\caption{\textbf{Ablation study.} All our contributions are important, removing the global spatial transformer leads to lower local details (bends legs), only global transformation does not correct shape and pose differences (thinner, straight legs), and predicting the vector field (not its derivative) produces foldovers. *silhouette estimated from an unpaired target domain image.}
	\label{fig:ablation}
\end{figure}

We compared our full model at PCK-15 (83.2), 
to not using one of our core contributions: no deformation (64.9), 
only global affine (57.3), 
only local non-linear (79.3), 
and directly encoding a vector field (69.0). 
The numbers and Fig.~\ref{fig:ablation} shows that all contributions are important.
Also end-to-end training with $D_I$ is important, as shown in Fig.~\ref{fig:ablation wo DI}, and by additional examples in the supplemental document.

\section{Limitations and future work}

For some domains the assumption of a target segmentation mask is constraining. For instance, our method is not applicable for transferring synthetic humans to real images on cluttered backgrounds. In the future, we plan on integrating unsupervised segmentation, as demonstrated by \cite{bielski2019emergence} for single-domain image generation.

Although we could synthesize a variety of poses for the worm and fish using a single stylized source image, our method was not able to synthesize entirely new \textit{Drosophila} poses, because crossing legs could not be modeled using a 2D image deformation. Therefore, sufficiently varied examples were needed in the source domain. 
Moreover, symmetries and self-similarities can lead to flipped limb identities. This remains a hard, open problem that we do not address in this study. Nevertheless, we believe that this is an important first step and that temporal cues and multi-view can be used to find a consistent and correct assignment in the future, following ideas used in \cite{yao2019monet} to perform pose estimation of humans and monkeys.

The attained accuracy is not yet perfect (see bottom of Fig.~\ref{fig:qualitative pose}) and can only be used to classify gross behaviours. Additional advances are needed to obtain pixel accurate reconstructions which would enable analysis of detailed movements, such as the activation of individual muscles.

\section{Conclusion}

In this paper, we have presented an approach for translating synthetic images to a real domain via explicit shape and pose deformation that consistently outperforms existing image translation methods.
Our method allows us to train a pose estimator on synthetic images that generalize to real ones; without requiring manual keypoint labels.
One of our test cases is on \textit{Drosophila} tethered to a microscope used to measure neural activity. By combining our improvements on pose estimation with state-of-the-art microscopy, we anticipate more rapid advances in understanding the relationship between animal behaviour and  neural activity.

\section{Acknowledgment}
PR acknowledges support from an SNSF Project grant (175667) and an SNSF Eccellenza grant (181239). PR and PF acknowledge support from an EPFL SV iPhD grant.

\bibliography{bib/egbib,bib/behav,bib/vision,bib/vision3,bib/misc,bib/short,bib/learning}

\begin{thebibliography}{10}\itemsep=-1pt

\bibitem{andriluka14}
Mykhaylo Andriluka, Leonid Pishchulin, Peter Gehler, and Bernt Schiele.
\newblock 2d human pose estimation: New benchmark and state of the art
  analysis.
\newblock In {\em Proceedings of the IEEE Conference on computer Vision and
  Pattern Recognition}, pages 3686--3693, 2014.

\bibitem{ashburner2007fast}
John Ashburner.
\newblock A fast diffeomorphic image registration algorithm.
\newblock {\em Neuroimage}, 38(1):95--113, 2007.

\bibitem{balakrishnan19}
Guha Balakrishnan, Amy Zhao, Mert Sabuncu, John Guttag, and Adrian~V. Dalca.
\newblock Voxelmorph: A learning framework for deformable medical image
  registration.
\newblock {\em IEEE TMI: Transactions on Medical Imaging}, 2019.

\bibitem{bielski2019emergence}
Adam Bielski and Paolo Favaro.
\newblock Emergence of object segmentation in perturbed generative models.
\newblock {\em arXiv preprint arXiv:1905.12663}, 2019.

\bibitem{bousmalis2017unsupervised}
Konstantinos Bousmalis, Nathan Silberman, David Dohan, Dumitru Erhan, and Dilip
  Krishnan.
\newblock Unsupervised pixel-level domain adaptation with generative
  adversarial networks.
\newblock In {\em CVPR}, volume~1, page~7, 2017.

\bibitem{chen2017photographic}
Qifeng Chen and Vladlen Koltun.
\newblock Photographic image synthesis with cascaded refinement networks.
\newblock In {\em ICCV}, volume~1, page~3, 2017.

\bibitem{dalca2019learning}
Adrian~V Dalca, Marianne Rakic, John Guttag, and Mert~R Sabuncu.
\newblock Learning conditional deformable templates with convolutional
  networks.
\newblock {\em NeurIPS}, 2019.

\bibitem{Engstrom2016}
Logan Engstrom.
\newblock Fast style transfer.
\newblock \url{https://github.com/lengstrom/fast-style-transfer/}, 2016.

\bibitem{Fu2019}
Huan Fu, Mingming Gong, Chaohui Wang, Kayhan Batmanghelich, Kun Zhang, and
  Dacheng Tao.
\newblock {Geometry-Consistent Generative Adversarial Networks for One-Sided
  Unsupervised Domain Mapping}.
\newblock In {\em CVPR}, 2019.

\bibitem{gatys2016preserving}
Leon~A Gatys, Matthias Bethge, Aaron Hertzmann, and Eli Shechtman.
\newblock Preserving color in neural artistic style transfer.
\newblock {\em arXiv preprint arXiv:1606.05897}, 2016.

\bibitem{gatys2015neural}
Leon~A Gatys, Alexander~S Ecker, and Matthias Bethge.
\newblock A neural algorithm of artistic style.
\newblock {\em arXiv preprint arXiv:1508.06576}, 2015.

\bibitem{gatys2016image}
Leon~A Gatys, Alexander~S Ecker, and Matthias Bethge.
\newblock Image style transfer using convolutional neural networks.
\newblock In {\em CVPR}, pages 2414--2423, 2016.

\bibitem{gokaslan2018improving}
Aaron Gokaslan, Vivek Ramanujan, Daniel Ritchie, Kwang In~Kim, and James
  Tompkin.
\newblock Improving shape deformation in unsupervised image-to-image
  translation.
\newblock In {\em Proceedings of the European Conference on Computer Vision
  (ECCV)}, pages 649--665, 2018.

\bibitem{Gokaslan18}
Aaron Gokaslan, Vivek Ramanujan, Daniel Ritchie, Kwang~In Kim, and James
  Tompkin.
\newblock Improving shape deformation in unsupervised image to image
  translation.
\newblock In {\em ECCV}, 2018.

\bibitem{Gunel19}
Semih Gunel, Helge Rhodin, Daniel Morales, João Compagnolo, Pavan Ramdya, and
  Pascal Fua.
\newblock Deepfly3d, a deep learning-based approach for 3d limb and appendage
  tracking in tethered, adult drosophila.
\newblock In {\em eLife}, 2019.

\bibitem{he2016dual}
Di He, Yingce Xia, Tao Qin, Liwei Wang, Nenghai Yu, Tie-Yan Liu, and Wei-Ying
  Ma.
\newblock Dual learning for machine translation.
\newblock In {\em NeurIPS}, pages 820--828, 2016.

\bibitem{Insafutdinov16}
Eldar Insafutdinov, Leonid Pishchulina, Bjoern Andres, Mykhaylo Andriluka, and
  Bernt Schiele.
\newblock Deepercut: A deeper, stronger, and faster multiperson pose estimation
  model.
\newblock In {\em ECCV}, 2016.

\bibitem{Ionescu14a}
Catalin Ionescu, Dragos Papava, Vlad Olaru, and Cristian Sminchisescu.
\newblock {{Human3.6M}: Large Scale Datasets and Predictive Methods for 3D
  Human Sensing in Natural Environments}.
\newblock 2014.

\bibitem{pix2pix2016}
Phillip Isola, Jun-Yan Zhu, Tinghui Zhou, and Alexei~A Efros.
\newblock Image-to-image translation with conditional adversarial networks.
\newblock {\em arxiv}, 2016.

\bibitem{isola2017image}
Phillip Isola, Jun-Yan Zhu, Tinghui Zhou, and Alexei~A Efros.
\newblock Image-to-image translation with conditional adversarial networks.
\newblock {\em CVPR}, 2017.

\bibitem{Jaderberg15}
Max Jaderberg, Karen Simonyan, Andrew Zisserman, and Koray Kavukcuoglu.
\newblock {Spatial Transformer Networks}.
\newblock pages 2017--2025, 2015.

\bibitem{Jing17}
Yongcheng Jing, Yezhou Yang, Zunlei Feng, Jingwen Ye, Yizhou Yu, and Mingli
  Song.
\newblock Neural style transfer: A review.
\newblock {\em arXiv}, 05 2017.

\bibitem{johnson2016perceptual}
Justin Johnson, Alexandre Alahi, and Li Fei-Fei.
\newblock Perceptual losses for real-time style transfer and super-resolution.
\newblock In {\em ECCV}, pages 694--711. Springer, 2016.

\bibitem{Johnson19}
Robert~Evan Johnson, Scott Linderman, Thomas Panier, Caroline~Lei Wee, Erin
  Song, Kristian~Joseph Herrera, Andrew Miller, and Florian Engert.
\newblock Probabilistic models of larval zebrafish behavior: Structure on many
  scales.
\newblock 2019.

\bibitem{kim2019unsupervised}
Boah Kim, Jieun Kim, June-Goo Lee, Dong~Hwan Kim, Seong~Ho Park, and Jong~Chul
  Ye.
\newblock Unsupervised deformable image registration using cycle-consistent
  cnn.
\newblock In {\em MICCAI}, pages 166--174. Springer, 2019.

\bibitem{kim2017learning}
Taeksoo Kim, Moonsu Cha, Hyunsoo Kim, Jung~Kwon Lee, and Jiwon Kim.
\newblock Learning to discover cross-domain relations with generative
  adversarial networks.
\newblock {\em ICML}, 2017.

\bibitem{li2016combining}
Chuan Li and Michael Wand.
\newblock Combining markov random fields and convolutional neural networks for
  image synthesis.
\newblock In {\em CVPR}, pages 2479--2486, 2016.

\bibitem{Lin14}
Tsung-Yi Lin, Michael Maire, Serge Belongie, James Hays, Pietro Perona, Deva
  Ramanan, Piotr Doll{\'a}r, and C.~Lawrence Zitnick.
\newblock Microsoft coco: Common objects in context.
\newblock In David Fleet, Tomas Pajdla, Bernt Schiele, and Tinne Tuytelaars,
  editors, {\em ECCV}, pages 740--755, Cham, 2014. Springer International
  Publishing.

\bibitem{liu2016coupled}
Ming-Yu Liu and Oncel Tuzel.
\newblock Coupled generative adversarial networks.
\newblock In {\em NeurIPS}, pages 469--477, 2016.

\bibitem{Mahmood19}
Naureen Mahmood, Nima Ghorbani, Nikolaus~F. Troje, Gerard Pons-Moll, and
  Michael~J. Black.
\newblock Amass: Archive of motion capture as surface shapes.
\newblock In {\em ICCV}, Oct 2019.

\bibitem{mechrez2018contextual}
Roey Mechrez, Itamar Talmi, and Lihi Zelnik-Manor.
\newblock The contextual loss for image transformation with non-aligned data.
\newblock {\em ECCV}, 2018.

\bibitem{Mehta17b}
Dushyant Mehta, Srinath Sridhar, Oleksandr Sotnychenko, Helge Rhodin, Mohammad
  Shafiei, Hans-Peter Seidel, Weipeng Xu, Dan Casas, and Christian Theoballt.
\newblock {Vnect: Real-Time 3D Human Pose Estimation with a Single RGB Camera}.
\newblock In {\em SIGGRAPH}, 2017.

\bibitem{Nath19}
Tanmay Nath, Alexander Mathis, An~Chi Chen, Amir Patel, Matthias Bethge, and
  Mackenzie~W Mathis.
\newblock Using deeplabcut for 3d markerless pose estimation across species and
  behaviors.
\newblock {\em Nature Protocols}, 2019.

\bibitem{Newell16}
Alejandro Newell, Kaiyu Yang, and Jia Deng.
\newblock {Stacked Hourglass Networks for Human Pose Estimation}.
\newblock {\em ECCV}, pages 483--499, 2016.

\bibitem{Pavlakos17}
Georgios Pavlakos, Xiaowei Zhou, Konstantinos~G. Derpanis, and Kostas
  Daniilidis.
\newblock {Harvesting Multiple Views for Marker-Less 3D Human Pose
  Annotations}.
\newblock In {\em CVPR}, 2017.

\bibitem{Pavllo19}
Dario Pavllo, Christoph Feichtenhofer, David Grangier, and Michael Auli.
\newblock 3d human pose estimation in video with temporal convolutions and
  semi-supervised training.
\newblock In {\em CVPR}, 2019.

\bibitem{Pereira18}
Talmo~D Pereira, Diego~E Aldarondo, Lindsay Willmore, Mikhail Kislin, Samuel
  S-H Wang, Mala Murthy, and Joshua~W Shaevitz.
\newblock Fast animal pose estimation using deep neural networks.
\newblock {\em Nature methods}, 16(1):117, 2019.

\bibitem{Popa17}
Alin-Ionut Popa, Mihai Zanfir, and Cristian Sminchisescu.
\newblock {Deep Multitask Architecture for Integrated 2D and 3D Human Sensing}.
\newblock In {\em CVPR}, 2017.

\bibitem{ramdya2017climbing}
Pavan Ramdya, Robin Thandiackal, Raphael Cherney, Thibault Asselborn, Richard
  Benton, Auke~Jan Ijspeert, and Dario Floreano.
\newblock Climbing favours the tripod gait over alternative faster insect
  gaits.
\newblock {\em Nature communications}, 8:14494, 2017.

\bibitem{Recasens18}
Adria Recasens, Petr Kellnhofer, Simon Stent, Wojciech Matusik, and Antonio
  Torralba.
\newblock Learning to zoom: a saliency-based sampling layer for neural
  networks.
\newblock In {\em ECCV}, pages 51--66, 2018.

\bibitem{Rhodin19a}
Helge Rhodin, Victor Constantin, Isinsu Katircioglu, Mateus Salzmann, and
  Pascal Fua.
\newblock {Neural Scene Decomposition for Human Motion Capture}.
\newblock In {\em CVPR}, 2019.

\bibitem{Rogez17}
Grégory Rogez, Philippe Weinzaepfel, and Cordelia Schmid.
\newblock {Lcr-Net: Localization-Classification-Regression for Human Pose}.
\newblock In {\em CVPR}, 2017.

\bibitem{Ronneberger15}
Olaf Ronneberger, Philipp Fischer, and Thomas Brox.
\newblock U-net: Convolutional networks for biomedical image segmentation.
\newblock In {\em International Conference on Medical image computing and
  computer-assisted intervention}, pages 234--241. Springer, 2015.

\bibitem{rueckert2006diffeomorphic}
Daniel Rueckert, Paul Aljabar, Rolf~A Heckemann, Joseph~V Hajnal, and Alexander
  Hammers.
\newblock Diffeomorphic registration using b-splines.
\newblock In {\em MICCAI}, pages 702--709. Springer, 2006.

\bibitem{sangkloy2017scribbler}
Patsorn Sangkloy, Jingwan Lu, Chen Fang, Fisher Yu, and James Hays.
\newblock Scribbler: Controlling deep image synthesis with sketch and color.
\newblock In {\em CVPR}, volume~2, 2017.

\bibitem{shrivastava2017learning}
Ashish Shrivastava, Tomas Pfister, Oncel Tuzel, Joshua Susskind, Wenda Wang,
  and Russell Webb.
\newblock Learning from simulated and unsupervised images through adversarial
  training.
\newblock In {\em CVPR}, volume~2, page~5, 2017.

\bibitem{Sigal10}
L. Sigal, A. Balan, and M.~J. Black.
\newblock {Humaneva: Synchronized Video and Motion Capture Dataset and Baseline
  Algorithm for Evaluation of Articulated Human Motion}.
\newblock 2010.

\bibitem{Sun17}
Xiao Sun, Jiaxiang Shang, Shuang Liang, and Yichen Wei.
\newblock {Compositional Human Pose Regression}.
\newblock In {\em ICCV}, 2017.

\bibitem{Balazs14}
Balazs Szigeti, Padraig Gleeson, Michael Vella, Sergey Khayrulin, Andrey
  Palyanov, Jim Hokanson, Michael Currie, Matteo Cantarelli, Giovanni Idili,
  and Stephen~D. Larson.
\newblock Openworm: an open-science approach to modeling caenorhabditis
  elegans.
\newblock {\em Front. Comput. Neurosci.}, 2014.

\bibitem{Tang18}
Wei Tang and Wu Ying.
\newblock Deeply learned compositional models for human pose estimation.
\newblock In {\em ECCV}, 2018.

\bibitem{ulyanov2016instance}
Dmitry Ulyanov, Andrea Vedaldi, and Victor Lempitsky.
\newblock Instance normalization: The missing ingredient for fast stylization.
\newblock {\em arXiv preprint arXiv:1607.08022}, 2016.

\bibitem{Varol17}
Gül Varol, Javier Romero, Xavier Martin, Naureen Mahmood, Michael~J. Black,
  and Ivan~Laptev an Cordelia~Schmid.
\newblock {Learning from Synthetic Humans}.
\newblock 2017.

\bibitem{Wandt_2019_CVPR}
Bastian Wandt and Bodo Rosenhahn.
\newblock Repnet: Weakly supervised training of an adversarial reprojection
  network for 3d human pose estimation.
\newblock In {\em CVPR}, June 2019.

\bibitem{Wang04imagequality}
Zhou Wang, Alan~Conrad Bovik, Hamid~Rahim Sheikh, Student Member, Eero~P.
  Simoncelli, and Senior Member.
\newblock Image quality assessment: From error visibility to structural
  similarity.
\newblock {\em IEEE Transactions on Image Processing}, 13:600--612, 2004.

\bibitem{Wayne19}
Wayne Wu, Kaidi Cao, Cheng Li, Chen Qian, and Chen~Change Loy.
\newblock Transgaga: Geometry-aware unsupervised image-to-image translation.
\newblock In {\em CVPR}, 2019.

\bibitem{Xiao18}
Bin Xiao, Haiping Wu, and Yichen Wei.
\newblock pages 466--481, 2018.

\bibitem{Xiu18}
Yuliang Xiu, Jiefeng Li, Haoyu Wang, Yinghong Fang, and Cewu Lu.
\newblock {Pose Flow}: Efficient online pose tracking.
\newblock In {\em BMVC}, 2018.

\bibitem{Yang17}
W. Yang, S. Li, W. Ouyang, and X. Wang.
\newblock Learning feature pyramids for human pose estimation.
\newblock In {\em ICCV}, 2017.

\bibitem{yao2019monet}
Yuan Yao, Yasamin Jafarian, and Hyun~Soo Park.
\newblock Monet: Multiview semi-supervised keypoint detection via epipolar
  divergence.
\newblock In {\em ICCV}, pages 753--762, 2019.

\bibitem{Yemini13}
Eviatar Yemini, Tadas Jucikas, Laura~J Grundy, André E~X Brown, and William~R
  Schafer.
\newblock A database of caenorhabditis elegans behavioral phenotypes.
\newblock In {\em Nature Methods}, 2013.

\bibitem{yin2018understanding}
Penghang Yin, Jiancheng Lyu, Shuai Zhang, Stanley~J. Osher, Yingyong Qi, and
  Jack Xin.
\newblock Understanding straight-through estimator in training activation
  quantized neural nets.
\newblock In {\em ICLR}, 2019.

\bibitem{Shu18}
Riza Alp Guler Dimitris Samaras Nikos~Paragios Zhixin~Shu, Mihir~Sahasrabudhe
  and Iasonas Kokkinos.
\newblock Deforming autoencoders: Unsupervised disentangling of shape and
  appearance.
\newblock In {\em ECCV}, 2018.

\bibitem{zhou2017towards}
Xingyi Zhou, Qixing Huang, Xiao Sun, Xiangyang Xue, and Yichen Wei.
\newblock Towards 3d human pose estimation in the wild: a weakly-supervised
  approach.
\newblock In {\em ICCV}, pages 398--407, 2017.

\bibitem{Zhou17d}
Xingyi Zhou, Qixing Huang, Xiao Sun, Xiangyang Xue, and Yichen Wei.
\newblock Weakly-supervised transfer for 3d human pose estimation in the wild.
\newblock In {\em IEEE International Conference on Computer Vision, ICCV},
  volume~3, page~7, 2017.

\bibitem{zhu2017unpaired}
Jun-Yan Zhu, Taesung Park, Phillip Isola, and Alexei~A Efros.
\newblock Unpaired image-to-image translation using cycle-consistent
  adversarial networks.
\newblock {\em ICCV}, 2017.

\bibitem{Zuffi19}
Silvia Zuffi, Angjoo Kanazawa, Tanya Berger-Wolf, and Michael~J. Black.
\newblock Three-d safari: Learning to estimate zebra pose, shape, and texture
  from images "in the wild".
\newblock In {\em ICCV}, 2019.

\end{thebibliography}
\bibliographystyle{ieee_fullname}


\section{Appendix}
In this document, we supply additional evaluation, training,
and implementation details, and provide a more details on the ablation study. The stability of the generated images is shown at hand of a short supplemental video.

\subsection{Additional qualitative results.}

\begin{figure*}[t]
	\begin{center}
		\includegraphics[width=1\linewidth]{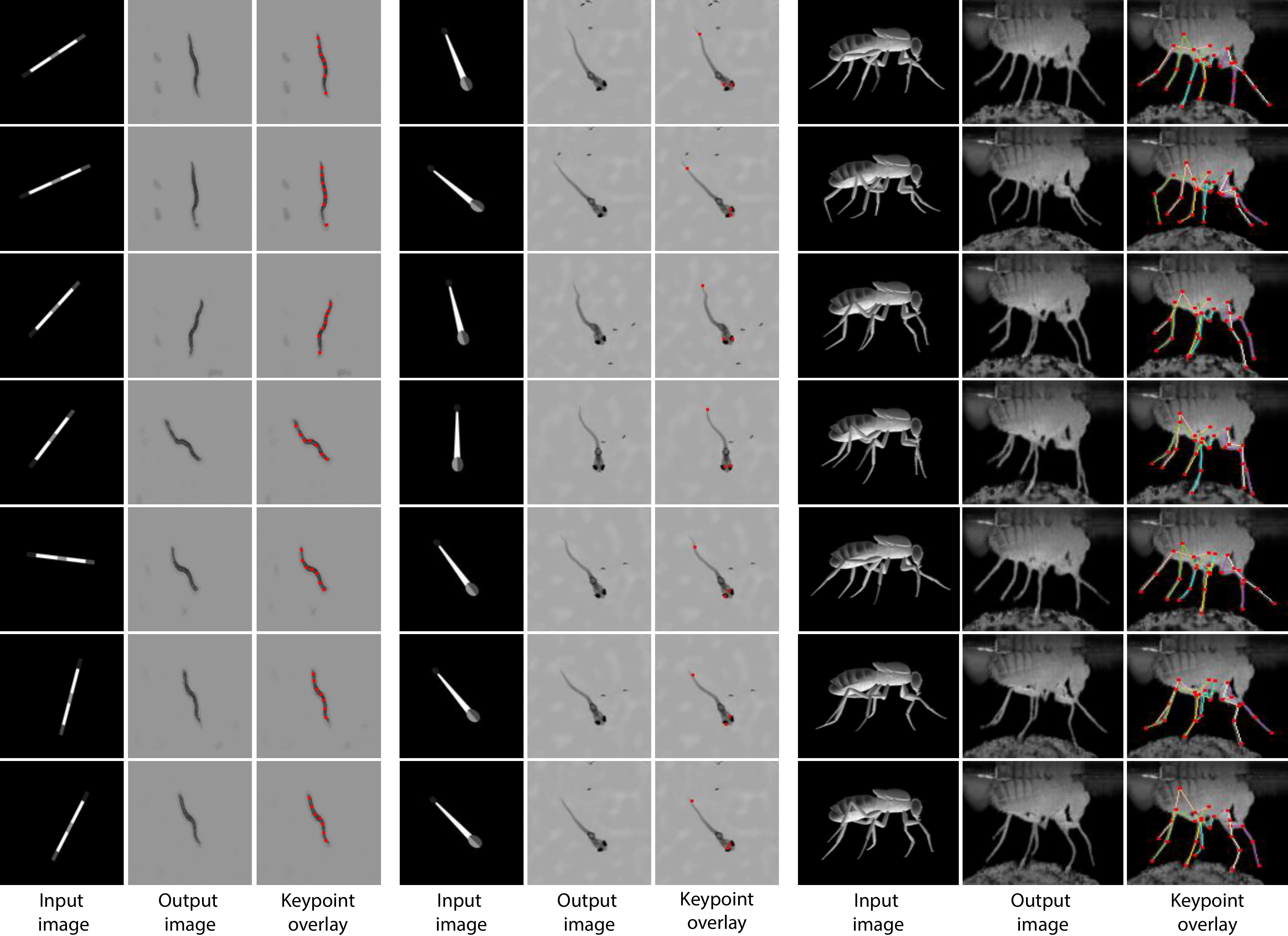}
	\end{center}
	\caption{\textbf{Qualitative image generation results.} Our approach can generate realistic and diverse poses, which are transferred across domains faithfully. Our method works on all three tested animals, including the \textit{Drosophila} dataset with superimposed legs that are on the ball that has no correspondence in the source domain.}
	\label{fig:gen all}
\end{figure*}

We included only few qualitative experiments in the main document due to space constraints.
Fig.~\ref{fig:gen all} provides additional examples of the image generation quality and the accuracy of the associated keypoint annotations, inferred via our explicit deformation field. 

\begin{figure*}[t]
	\begin{center}
		\includegraphics[width=1\linewidth]{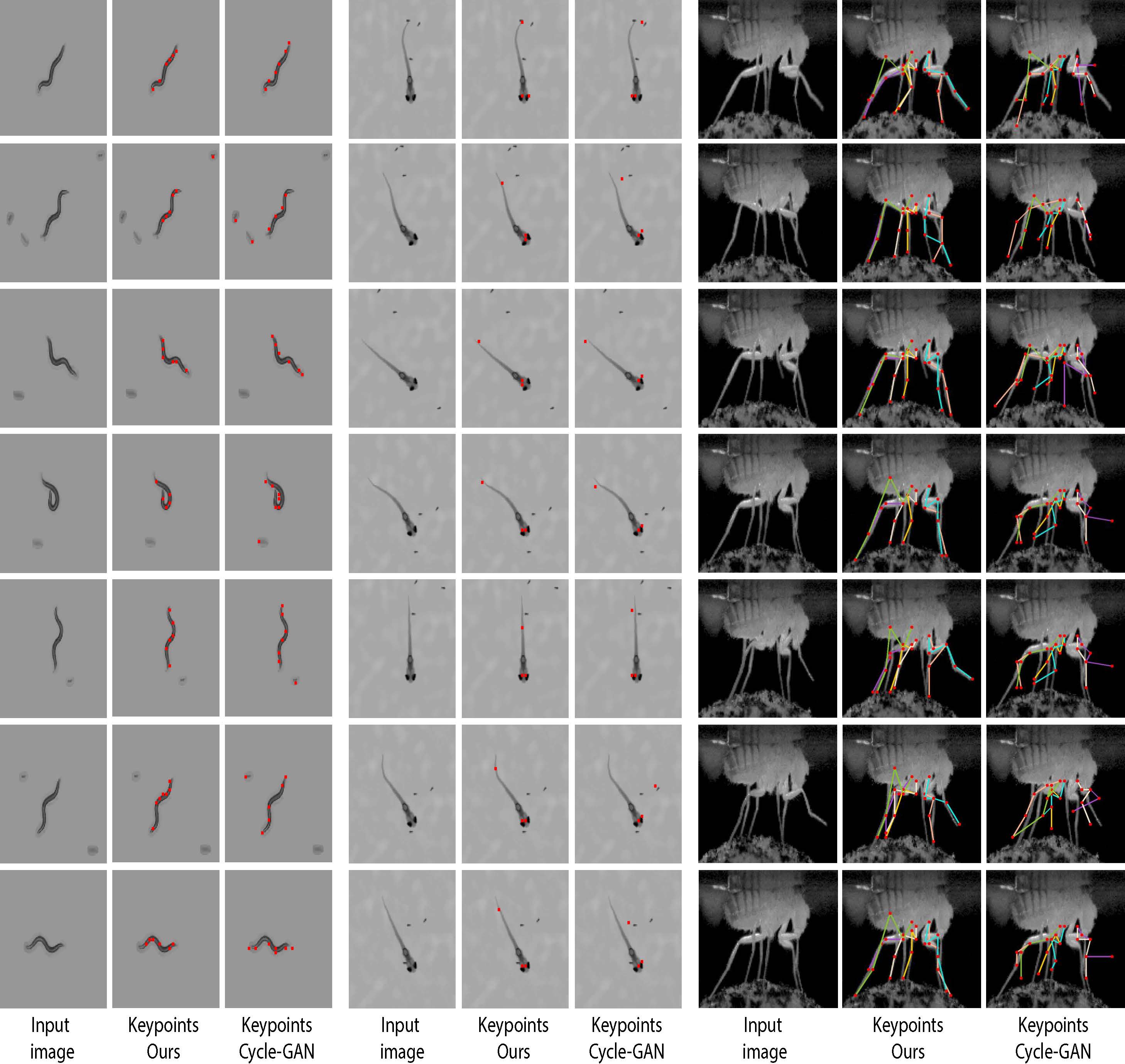}
	\end{center}
	\caption{\textbf{Pose estimation result comparison.} Training a pose estimator on our generated images yields accurate detections with far less failures when compared to Cycle-GAN, the best performing baseline. The \textit{Drosophila} case is most challenging as the legs are thin and self-similar.}
	\label{fig:pose all}
\end{figure*}

Moreover, Fig.~\ref{fig:pose all} shows additional examples of the pose estimation quality compared to using Cycle-GAN. Our approach produces much fewer miss classifications, for instance, in the case of extreme bending positions of the worm.

\subsection{Ablation study details.}

The ablation study in the main document tests our complete approach while removing of our core contributions in terms of the PCK metric at threshold 15 pixels. The additional metrics in Table~\ref{tab:ablation} show that our contributions improve consistently across different PCK thresholds. 

Each of our contributions is significant with gains of 8 to 30 on PCK-15 and 4 to 14 AUC points.
Notably, using global affine deformation is worse than without any deformation. This may be because the affine network rotates the body of synthetic fly to match the shape of real fly. However, the rotation also affects the leg orientation, which leads to less realistic poses.
It is best to use global and local motion together (Ours).

\begin{figure}[t]
	\begin{center}
		\includegraphics[width=1\linewidth]{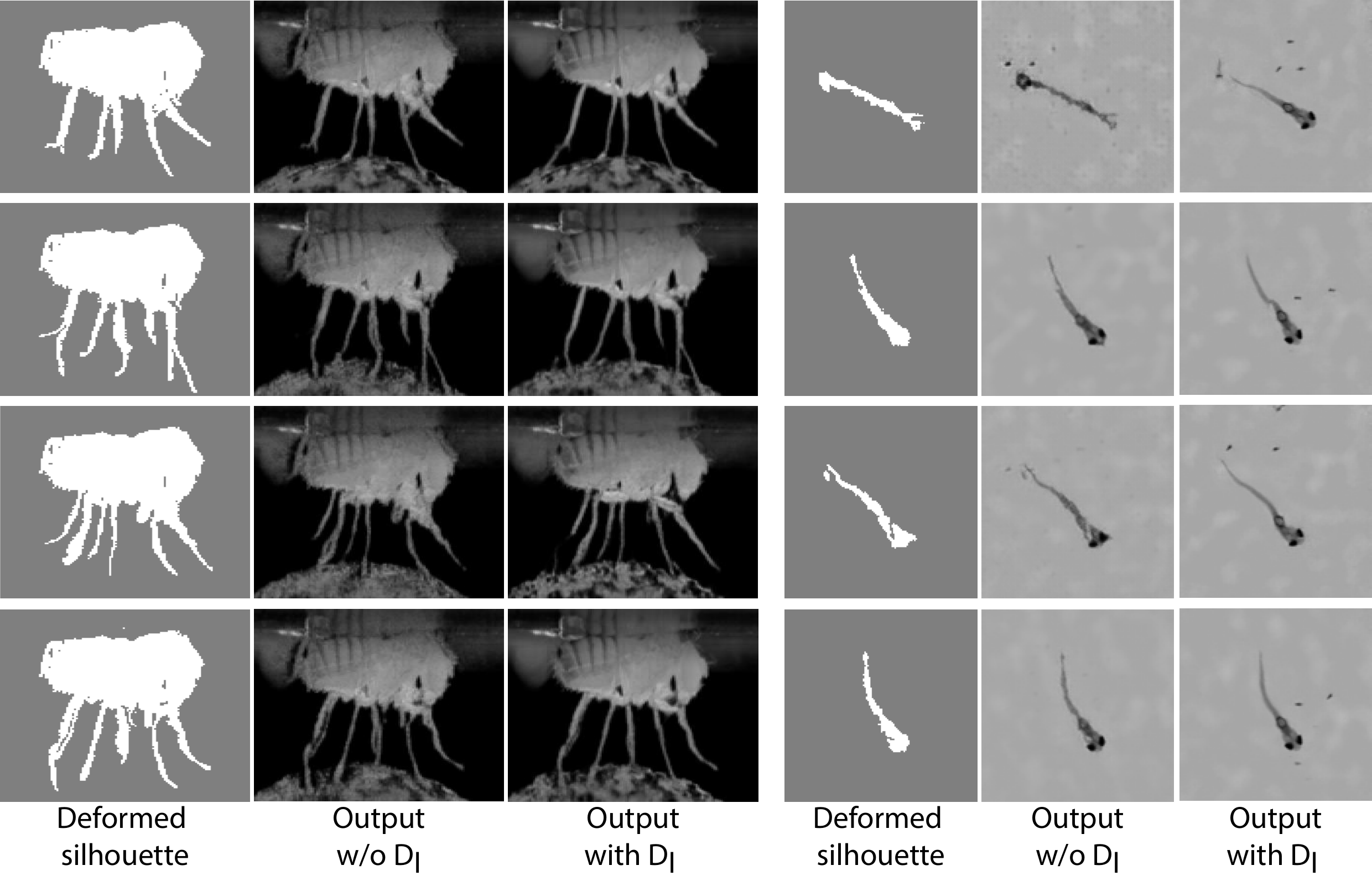}
	\end{center}
	\caption{\textbf{Ablation study on $D_I$.} Without $D_I$, small artifacts in the generated (deformed) segmentation masks lead to unrealistic images.}
	\label{fig:more DI}
\end{figure}

Moreover, Fig.~\ref{fig:more DI} provides additional results comparing the generated image quality with and without using $D_I$. Clear improvements are gained for the fish and Drosophila. For instance, legs are properly superimposed on the ball, while holes arise without $D_I$ (therefore, without end-to-end training). No significant improvement could be observed on the  worm case due to its simplicity.

\begin{table}[]
	\centering
	\resizebox{1\columnwidth}{!}
	{
		\begin{tabular}{|l|c|ccc|}
			\hline
			\multicolumn{1}{|l|}{Metric} & \multicolumn{1}{|l|}{\begin{tabular}[c]{@{}c@{}}Batch\\size\end{tabular}} & \multicolumn{1}{l|}{\begin{tabular}[c]{@{}l@{}}PI-PCK $\uparrow$\\ (5 pix)\end{tabular}} & \multicolumn{1}{l|}{\begin{tabular}[c]{@{}l@{}}PI-PCK $\uparrow$\\ (15 pix)\end{tabular}} & \multicolumn{1}{l|}{\begin{tabular}[c]{@{}l@{}}PI-AUC $\uparrow$\\ (4-45 pix)\end{tabular}}
			\\ \hline
			Ours & 128 & {28.0} & {74.4} & {80.6} \\
			\hline
			Ours & 12 & \textbf{38.6} & \textbf{83.2} & \textbf{85.1} \\
			Ours w/o global deformation & 12 & {31.4} & {79.2} & {84.0} \\
			Ours w/o deformation & 12 & {18.5} & {64.9} & {74.9} \\
			Ours w/o local deformation & 12 & {13.1} & {57.4} & {73.8} \\
			Ours using vector field & 12 & {18.6} & {69.1} & {79.0} \\
			\hline
		\end{tabular}
		
	}
	\smallskip
	\caption{\textbf{Detailed ablation study on \textit{Drosophila Melanogaster}.} All model components contribute to the final reconstruction accuracy. Surprisingly, a smaller batch size improved results.}
	\label{tab:ablation}
\end{table}

\subsection{Dataset sources and splits.}

The worm dataset stems from the OpenWorm initiative \cite{Yemini13,Balazs14}. We used three videos after subsampling to 8x speed. The OpenWorm videos are referred by strain type and timestamp. We used the three videos specified in Table~\ref{tab:worm specs}, downloaded from YouTube at subsampled framerate (8x speed compared to the original recording).\\
\begin{table}[h!]
	\centering
\resizebox{1\columnwidth}{!}
{
\begin{tabular}{c|c|c}
Strain & Strain description & Time stamp\\
\hline
OW940 & zgIs128[P(dat-1)::alpha-Synuclein::YFP] & 2014-03-14T13:39:36+01:00 \\
OW940 & zgIs128[P(dat-1)::alpha-Synuclein::YFP] & 2014-03-06T09:11:51+01:00 \\
OW939 & zgIs113[P(dat-1)::alpha-Synuclein::YFP] & 2014-02-22T14:13:49+01:00 \\
\end{tabular}
}
	\smallskip
\caption{\textbf{OpenWorm videos.} Strain type and timestamp of the used videos published by \cite{Yemini13,Balazs14}.}
\label{tab:worm specs}
\end{table}

The worm is tracked in each video to be roughly centered. The only transformation done is scaling the original frames to resolution $128\times128$ pixels. We randomly picked 100 frames of these three videos for test and then picked 1000 frames out of all remaining frames for unpaired training. We manually annotated every 10th frame (100 frames) from the unpaired training examples with two keypoints (head and tail) to train the supervised baseline, and the entire test set (100 frames) for quantifying pose estimation accuracy.

For the zebrafish larva experiments, we used Video~3 (\url{672246_file04.avi}) published in the supplemental of \cite{Johnson19} (\url{biorxiv.org/content/10.1101/672246v1.supplementary-material}). We crop the original video from $1920\times1080$ pixels to the region with top left corner $(500,10)$ and bottom right $(1290,800)$, and scale it to $128\times128$ pixels. We deleted some repetitive frames where the zebrafish is not moving to increase the percentage of frames where zebrafish is bending. In total, we retained 600 frames. We selected the last 100 frames for test and 500 left for unpaired training. Besides the test images, we also manually annotated every 5th (100 frames) from the 500 training images as the training data for the supervised baseline.

\subsection{Training details}
\vspace{0.1cm}
\parag{Training The Unpaired Image Translation Network.}
We use the Adam optimizer with different initial learning rates for different modules. For $G_I$, $D_I$, we set the learning rate to $2\mathrm{e}{-3}$. For $G_S$, we set the learning rate to $2\mathrm{e}{-5}$ since a slight update will have a big impact on the deformation field due to the integrating the spatial gradient in the last layer of $G_S$. We set the learning rate of $D_S$ to $1/10$ the one of $G_S$, which balanced the influence of $G_S$ and $D_S$ in our experiments. 
In case of  \textit{Drosophila} training, we apply linear decay to our learning rates. We start the decay of $G_S$, $D_S$ at epoch 50 and reduce it to $0$ till epoch 100. 
For fish and worm, we set the learning rate of $G_D$ to $1e-4$ and $D_S$ to $1e-5$, to account for the simpler setting of deforming from a single template image. Moreover, we linearly decay from epoch 100 to epoch 200.

The batch size of the image translation training is set to 4. 
An other important detail is the initialization of $G_S$ to generate the identity mapping. We achieved that by initially training $G_S$ solely on the regularization term, which pushes it towards this state.

\parag{Training Pose Estimation Network.}
We use Adam optimizer with initial learning rate of $2\mathrm{e}{-3}$. We train the pose estimation network for 200 epochs and the learning rate starts to linear decay after epoch 100, till epoch 200.

\comment{
\begin{table}[]
	\centering
	\resizebox{0.9\columnwidth}{!}
	{
		\begin{tabular}{|l|cccc|}
			\hline
			&\multicolumn{4}{|c|}{\it Drosophila Melanogaster} \\ \hline
			\multicolumn{1}{|l|}{Metric} & \multicolumn{1}{l|}{\begin{tabular}[c]{@{}l@{}}PI-PCK $\uparrow$\\ (5 pix)\end{tabular}} & \multicolumn{1}{l|}{\begin{tabular}[c]{@{}l@{}}PI-PCK $\uparrow$\\ (15 pix)\end{tabular}} & \multicolumn{1}{l|}{\begin{tabular}[c]{@{}l@{}}PI-AUC $\uparrow$\\ (4-45 pix)\end{tabular}} & \begin{tabular}[c]{@{}l@{}}PI-RMSE $\downarrow$\\ (pix)\end{tabular} \\ \hline
			Synthetic & 22.0 & 61.8 & 72.9 & 15.040 \\
			Fast-Style-Transfer & 11.1 & 49.1 & 63.0 & 19.856 \\
			Gc-GAN & 11.6 & 65.2 & 75.1 & 13.676 \\
			Cycle-GAN & 14.3 & 75.3 & 77.5 & 12.701 \\
			Ours & \textbf {37.5} & \textbf{82.0} & \textbf{84.6} & \textbf{9.455} \\
			\hline
			Supervised & 72.7 & 89.0 & 90.7 & 6.293 \\
			\hline
		\end{tabular}
	}
	\smallskip
	\caption{\textbf{Pose estimation accuracy comparison on \textit{Drosophila Melanogaster}.} A similar improvement as for Drosophila is attained on the other tested laboratory animals, with a particularly big improvements on the zebrafish. Pose-invariant training improves results.}
	\label{tab:poseD}
\end{table}
}

\subsection{Implementation details}
\label{sec:implementation}

\vspace{0.1cm}
\parag{Deformation representation.} Directly modeling the deformation as vector field will make the transformation unstable and easily lose the semantic correspondence.  For example, a vector field permits coordinate crossing and disconnected areas, which leads to unstable training and divergence. In order to preserve a connected grid topology, we model our deformation close to the difformorphic transformation, which generates the deformation field as the integral of a velocity field. This leads to useful properties such as invertibility and none crossing intersections~\cite{ashburner2007fast}.  
However, it is in general expensive to compute the integral over an axis, thus making it difficult to incorporate into deep networks. Instead of modeling a continuous velocity function, we directly model our deformation field $\phi$ as the integral of the spatial gradient of vector field, as proposed by Shu et al. \cite{Shu18}. We write,
\begin{equation}
\begin{aligned}
& \bigtriangledown\phi_x  = \frac{\partial \phi}{\partial x}  \  \ \ \ \ \ \  \ \bigtriangledown \phi_y = \frac{\partial \phi}{\partial y} 
\end{aligned}  
\end{equation}
where $x$, $y$ define the gradient directions along the image axes.  The $\phi_x$ and $\phi_y$ measure the difference of consecutive pixels.  By enforcing the difference to be positive (e.g., by using ReLU activation functions; we use HardTanh with range (0, 0.1)), we avoid self-crossing and unwanted disconnected areas. For example, when $\phi_x$ and $\phi_y$ equals to 1, the distance between the consecutive pixels is the the same. If  $\phi_x, \phi_y > 1$ , the distance will increase, otherwise, when  $\phi_x$ ,$\phi_y< 1$,  it will decrease.

The second module is the spatial integral layer, also the last layer of deformation spatial gradient generator. This layer sums the spatial gradients along the x and y directions and produces the final deformation field,
\begin{equation}
\phi_{i,j} = (\sum_{m = 0}^{i} \bigtriangledown \phi_{x_{m}} , \ \ \sum_{n = 0}^{j}\bigtriangledown \phi_{y_{n}}),
\end{equation}
where $i,j$ is the pixel location. 
\comment{
We formulate the pixel-wise transformation under this deformation field as 
\begin{equation}
\binom{u}{v} = \phi\binom{x}{y} = \binom{x+ dx}{y + dy}
\end{equation}
where $u$, $v$ are the source coordinates generated by the deformation field. For instance, for any single pixel located at (x, y) in the deformed image, the sampling location from the source image should be $u$, $v$.}
Since the $u$, $v$ in general position do not correspond to one exact pixel location in the source image, we compute the output image using a differentiable bilinear interpolation operation, as for spatial transformers \cite{Jaderberg15}.

\comment{
\begin{equation}
I(x,y) = \sum_{j}^{w}\sum_{i}^{h}I_{ori}(i, j)max(0, 1 - |u - i|)max(1 - |v-j|)
\end{equation}
The fig shows the deformation process.  

Note that we use the masks to represent the shape, thus $A_s$ means the mask of source image, $B_s$ means the mask of target image.
Based on generated deformation field, we directly sample from the original mask image $A_{s}$ and generate a new deformed mask $A_{d}$ .  
}

\parag{Shape Discriminator}

We utilize the $70 \times 70$ patchGAN discriminator as our backbone structure~\cite{pix2pix2016} . The patch-wise design makes the network focus on the local area of the shape. Furthermore, if the shape between two domains is extremely different, the patch-wise design prevents the discriminator from converging too quickly. However, the design also limits the network’s awareness of global shape changes \cite{gokaslan2018improving}. Thus, we add dilation to the second and the third convolution layers of patchGAN. Those dilated layers enlarge the receptive field of our shape discriminator, making it aware of bigger shape variation, giving a better guidance to the generator. 

\parag{Image Generator.}
We build our generator on the U-Net architecture, which is proved to be effective in tasks such as pixel-wise image translation and segmentation~\cite{Ronneberger15}. The generator contains several fully convolutional down-sampling and up-sampling layers. The skip connections in the generator help to propagate information directly from input features to the output, which guarantee the preservation of spatial information in the output image.

\parag{Pose Estimator.}

We adopt the stacked hourglass human pose estimation network to perform pose estimation on animals~\cite{Newell16}. The stacked hourglass network contains several repeated bottom-up, top-down processing modules with intermediate supervision between them. A single stack hourglass module consists of several residual bottleneck layers with max-pooling, following by the up-sampling layers and skip connections. We used 2 hourglass modules in our experiments. The pose estimation network is trained purely on the animal data we generated; without pre-training and manually annotated labels. The ground-truth poses come from the annotations of synthetic animal models. The pose invariant (PI) training is performed in all experiments labeled with \emph{PI training}.

\parag{Pose annotation.}
 
\textit{Drosophila} has six limbs, each limb has five joints, giving 30 2D keypoints that we aim to detect.  By using our image translation model, we generated $1500$ images with annotation from the synthetic data. Each image is in size $128 \times 128$ pixels. The first hourglass network is preceded with convolutional layers that reduce the input image size from $128 \times 128$ to $32 \times 32$. The second hourglass does not change the dimension. Thus, the network will output a $30\times 32 \times 32$ tensor, which represents the probability maps of 30 different joints locations. For training, we create the ground truth label using a 2D Gaussian with mean at the annotated keypoint and 0.5 on the diagonal of the covariance matrix. The training loss is the MSE between the generated probability map and the ground truth label.

We annotated three keypoints on D. rerio and seven keypoints on C. elegans. We use the same network as for Drosophila, but the output tensor adapted to the number of keypoints, $3\times 32 \times 32$ and $7\times 32 \times 32$, respectively.

\end{document}